\documentclass[times,twocolumn,final]{elsarticle}
\usepackage{graphicx} 
\usepackage{subcaption} 
\usepackage{medima}
\usepackage{framed,multirow}
\usepackage{subfiles}
\usepackage{amssymb}
\usepackage{amsmath}
\usepackage{latexsym}
\usepackage{bm}
\usepackage{xcolor}
\usepackage{ulem}
\usepackage{url}
\usepackage{tabularx}
\usepackage{arydshln}
\usepackage{booktabs}
\usepackage{hyperref}
\usepackage{makecell}

\newtheorem{definition}{Definition}
\newtheorem{proposition}{Proposition}

\journal{Medical Image Analysis}

\begin{document}

\verso{Y. Liu \textit{et~al.}}

\begin{frontmatter}

\title{MERIT: Multi-view Evidential Learning for Reliable and Interpretable Liver Fibrosis Staging \tnoteref{funding}}%
\tnotetext[funding]{This work was supported by the National Natural Science Foundation of China (grant No. 62372115, 62201014, and 62111530195), the Clinical Medicine Plus X-Young Scholars Project of Peking University (PKU2024LCXQ028), and the Fundamental Research Funds for the Central Universities.}

\author[1]{Yuanye \snm{Liu}\fnref{equal}}
\author[1]{Zheyao \snm{Gao}\fnref{equal}}
\author[2]{Nannan \snm{Shi}\fnref{equal}}
\fntext[equal]{These authors contributed equally to this work.}
\author[3]{Fuping \snm{Wu}}
\author[2]{Yuxin \snm{Shi}\corref{*}}
\author[4,5,6]{Qingchao \snm{Chen}\corref{*}}
\author[1]{Xiahai \snm{Zhuang}\corref{*}}
\cortext[*]{Corresponding author e-mail: zxh@fudan.edu.cn (Xiahai Zhuang), qingchao.chen@pku.edu.cn (Qingchao Chen), shiyuxin@shaphc.org (Yuxin Shi)}

\address[1]{School of Data Science, Fudan University, Shanghai, 200433, China}
\address[2]{Department of Radiology, Shanghai Public Health Clinical Center, Fudan University, Shanghai, 201508, China}
\address[3]{Nuﬃeld Department of Population Health, University of Oxford, Oxford, UK}
\address[4]{National Institute of Health Data Science, Peking University, Beijing, 100191, China}
\address[5]{Institute of Medical Technology, Peking University, Beijing, 100191, China}
\address[6]{State Key Laboratory of General Artificial Intelligence, Peking University, Beijing, 100191, China}

\received{24 July 2024}
\finalform{31 December 2024}
\accepted{11 February 2025}
\availableonline{**}
\communicated{**}

\begin{abstract}
Accurate staging of liver fibrosis from magnetic resonance imaging (MRI) is crucial in clinical practice. While conventional methods often focus on a specific sub-region, multi-view learning captures more information by analyzing multiple patches simultaneously. 
However, previous multi-view approaches could not typically calculate uncertainty by nature, and they generally integrate features from different views in a black-box fashion, hence compromising reliability as well as interpretability of the resulting models. 
In this work, we propose a new multi-view method based on evidential learning, referred to as MERIT, which tackles the two challenges in a unified framework. 
MERIT enables uncertainty quantification of the predictions to enhance reliability, and employs a logic-based combination rule to improve interpretability. 
Specifically, MERIT models the prediction from each sub-view as an opinion with quantified uncertainty under the guidance of the subjective logic theory. 
Furthermore, a distribution-aware base rate is introduced to enhance performance, particularly in scenarios involving class distribution shifts. 
Finally, MERIT adopts a feature-specific combination rule to explicitly fuse multi-view predictions, thereby enhancing interpretability. 
Results have showcased the effectiveness of the proposed MERIT, highlighting the reliability and offering both ad-hoc and post-hoc interpretability. They also illustrate that MERIT can elucidate the significance of each view in the decision-making process for liver fibrosis staging. Our code will be released via \href{https://github.com/HenryLau7/MERIT}{https://github.com/HenryLau7/MERIT}.
\end{abstract}


\end{frontmatter}


\section{Introduction}

\begin{figure*}[t]
    \centering 
    \begin{tabular}{cc}
        \includegraphics[width=1.0\columnwidth]{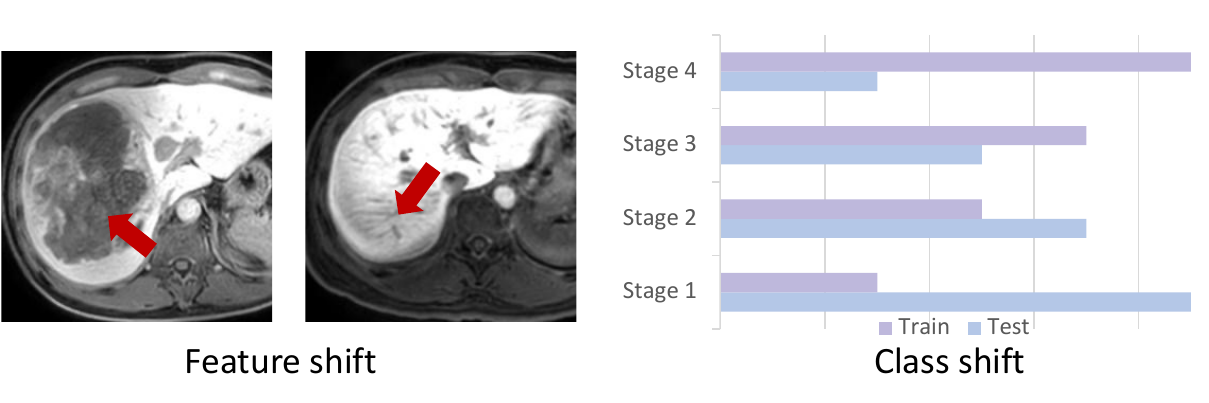}&
        \includegraphics[width=0.9\columnwidth]{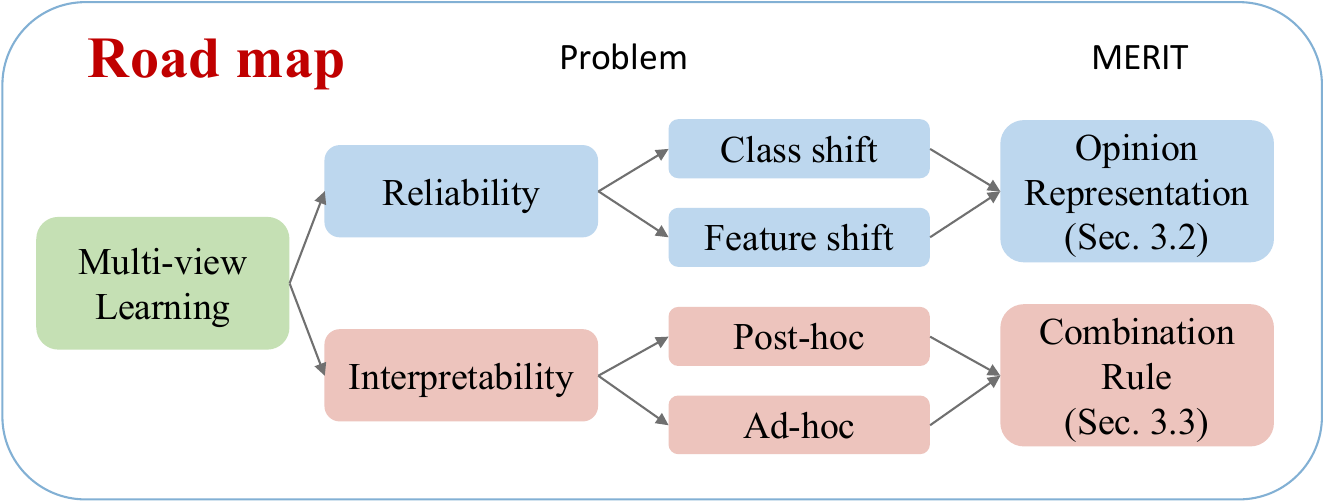}\\
         (a)  & (b) \\
\end{tabular}
\caption{(a) Distribution shift: The feature distribution shift in our problem is mainly caused by low-quality images with liver mass occupation (left) or artifacts (right). The class distribution shift denotes the difference in class proportion between training and test data.
(b) Roadmap: Our MERIT framework improves the reliability and interpretability of multi-view learning through opinion representation and belief fusion operators.}
\label{fig:shift_roadmap}
\end{figure*}

Liver fibrosis results from excessive accumulation of extracellular matrix proteins, particularly collagen, which may progress to liver cirrhosis and even lead to liver cancer \citep{J_2022AR_Liver_Shi}.
The Scheuer system serves as a popular standard for assessing the stage of fibrosis, categorizing it into four distinct stages (S1-S4), with S1 indicating fibrous portal expansion and S4 representing severe cirrhosis \citep{J_1991_scheuer}.
Accurate staging of fibrosis severity holds great importance in the diagnosis and treatment planning for various types of chronic liver diseases. 
Previous studies have identified specific MRI features that can aid in detecting liver fibrosis, with the use of Gd-EOB-DTPA enhancing the assessment \citep{J_2019_CT_liver, J_2022_pmx}.

Deep learning-based methods have emerged as a powerful framework for medical image analysis in recent studies \citep{J_2017MIA_medical_als_survey,B_2021_medical_cls_survey}. 
Various methods have been proposed for liver fibrosis staging, including convolutional neural networks (CNNs) \citep{J_2018R_pmx_liver,J_2020_liver_resnet}, contrast learning \citep{J_2019_liver_contrast},  transfer learning \citep{J_2020ER_liver_transfer,J_2021ER_liver_transfer} and advanced GAN \citep{J_2022_Liver_gan}. 
However, these approaches used the entire liver scan as input for training the model, which can lead to inadequate feature extraction due to the irregular shape of the liver and the presence of unrelated anatomical structures within the abdominal image. 
Furthermore, considering that the signs of liver fibrosis, such as changes in surface nodularity, widening of fissures, and imbalance between the left and right lobes, are distributed throughout the whole liver \citep{J_2002_fibrosis_feature_1,J_2015_fibrosis_feature_2,J_2021_fibrosis_feature_3}, capturing intricate fibrosis features proves challenging when utilizing the entire scan as a single input. 
Recently, \citet{J_2021ER_subview_liver} proposed using a sliding window to crop an image patch for data augmentation. However, they utilized only one patch as input at a time, capturing just a sub-view of the liver. 
To exploit informative features across the entire liver, we formulate this task as a multi-view learning problem, exploiting local and global features simultaneously.

Multi-view learning aims to exploit complementary information from multiple features \citep{J_2018PAMI_multi-view_zhang, J_2021_multiview_review, J_2022_trusted_mv}. 
The primary challenge lies in integrating features from multiple views properly. 
Decision-level fusion is frequently preferred for its ability to enhance the analytics of large data through systematic design and efficient process while providing better and unbiased results \citep{J_2010_decision_fusion_pro_1,J_2020_decision_fusion_pro_2}. 
Most decision-level fusion methods utilize aggregating strategies, such as simply averaging \citep{C_2014NIPS_decision_fusion_1, J_2021_decision_fusion_averaging}, majority voting \citep{J_2014_decision_fusion_voting,J_2024_decision_fusion_voting_2}, or learnable models \citep{C_2019ICCV_decision_fusion_learn, C_2022_devicion_fusion_learn_2}. 
However, in these methods, the weight of multi-view features is typically either fixed, limiting the model’s flexibility, or learned implicitly through model training, which compromises the interpretability of the decision-making process. 
More importantly, these methods are not capable of quantifying uncertainties, which is crucial for ensuring trustworthiness in healthcare applications. 

To enhance the reliability of multi-view learning, previous methods mainly apply Monte-Carlo (MC) dropout \citep{C_2016ICML_dropout}, variational inference (VI) \citep{C_2019CVPR_Uncertainty_VI}, or ensemble \citep{C_2021CVPR_masksembles} to involve model uncertainty (also called epistemic uncertainty) \citep{C_2017NIPS_uncertain}.
However, such models are not constructed to explicitly capture uncertainties due to mismatch between the
distributions of test and training data.
PriorNet \citep{C_2018NIPS_prior_net} first introduces distributional uncertainty by endowing a prior Dirichlet distribution to the classification model, which enables the detection of out-of-distribution (OOD) samples.
Evidential learning \citep{C_2018NIPS_Evidential,C_2023ICML_evidential_medical_2,C_2024CVPR_evidential_medical_1} provides an alternative way to model the distributional uncertainty under the guidance of evidence theory \citep{J_1968_evidence_theory}. 
In evidential learning, the output of a neural network is represented as an opinion that could quantify distributional uncertainty induced by the Dirichlet distribution.
However, the methods above mainly focus on the uncertainty caused by the feature distribution shift, neglecting the impact of the class distribution shift.
As shown in Fig. \ref{fig:shift_roadmap}(a), both feature and class distribution shifts could exist in the liver fibrosis staging task.
Thus, their predictions could be unreliable in such cases.

Following evidential learning, this work introduces MERIT, an evidential multi-view learning framework for liver fibrosis staging. 
As shown in Fig. \ref{fig:shift_roadmap}(b), it enhances reliability in distribution shift scenarios by representing neural network outputs as opinions based on the subjective logic theory \citep{B_2016_subjective}. 
Moreover, MERIT integrates opinions from multiple sources using feature-specific belief fusion operators, enabling both model-based (or ad-hoc) and post-hoc interpretability \citep{J_2019_interpret}. 
The ad-hoc interpretability comes from the modular design of our model, where the decision-making process based on opinions is interpretable and the black-box neural network only serves as a tool to support opinion representation. 
For the post-hoc interpretability, the trained model could provide the prediction-level interpretation. The contribution of each component of the feature, \textit{i.e.}, sub-view of the liver, could be measured through uncertainties.

This work is extended from our conference paper presented at MICCAI 2023 \citep{C_2023MICCAI_ours},
in which we have proposed an uncertainty-aware multi-view learning method with an interpretable fusion strategy for liver ﬁbrosis staging. 
In addition to the previous work, the novel contributions in this paper are listed as follows:
\begin{itemize}
    \item We incorporate a class distribution-aware base rate into evidential deep learning to address the issue of class distribution shift.
    \item We improve the combination rule by introducing feature-specific fusion operators, which faithfully model the relationship between local and global view features. Additionally, we provide a theoretical analysis of its applicability in our task to elaborate on its interpretation.
    \item We conduct extensive experiments on an extended multi-center liver MRI dataset, including the evaluation of reliability in feature and class distribution shift scenarios, as well as feature ablation study for post-hoc interpretability. 
\end{itemize}

The remainder of the article is organized as follows. 
Section \ref{sec:related_work} discusses the related work to this study. 
Section \ref{sec:method} elucidates the proposed MERIT framework for liver fibrosis staging. 
Section \ref{sec:experiment} introduces the experimental setups and evaluation results of our method on reliability, interpretability and multi-view modeling. Section \ref{sec:conclusion} concludes the study.

\section{Related works}\label{sec:related_work}
This work is mainly related to four research areas, \textit{i.e.},  liver fibrosis staging, multi-view learning, uncertainty quantification and evidence theory. 

\subsection{Deep Learning Methods for Liver Fibrosis Staging}
Accurate and timely staging of liver fibrosis is essential for the effective diagnosis and treatment planning for liver disease. 
Deep learning algorithms have shown advanced performance for liver fibrosis staging. Convolutional networks, such as 8-layer network \citep{J_2018ER_liver_CT_DL} and ResNet\citep{J_2020_liver_resnet} were utilized to stage liver fibrosis. \citet{J_2019_liver_contrast} employed a tailored CNN to extract features and contrast the similarity of different feature maps by applying the ModIBS measurement technique. Furthermore, \citet{J_2020ER_liver_transfer} and \citet{J_2021ER_liver_transfer} leveraged networks pre-trained on ImageNet to improve feature extraction for liver fibrosis staging. 
\citet{J_2022_Liver_gan} applied generative adversarial network (GAN) for data augmentation to improve the performance of liver fibrosis staging.
\citet{J_2022_Liver_dual} modeled the measurement of liver fibrosis and inflammation as a dual-task challenge, subsequently proposing a dual-task CNN to address these tasks effectively.

\subsection{Multi-view Fusion}
In recent years, multi-view learning has emerged as a potent method for extracting complementary features from multiple data sources to construct a comprehensive final prediction model, where the integration of predictions from different views is crucial. Beyond basic methods that merely concatenated features or summed them up at input level \citep{C_2016CVPR_input_fusion}, advanced feature-level fusion strategies have been developed. These strategies aimed to find a common representation across different views through canonical correlation analysis \citep{J_2002_fibrosis_feature_1,J_2015_fibrosis_feature_2} or maximizing the mutual information between different views using contrastive learning \citep{C_2020ECCV_Contrastive,C_2020ICML_Contrastive}.
In terms of decision-level fusion, common methods include decision averaging \citep{J_2018PAMI_decision_fusion_2}, decision voting \citep{J_2014_decision_fusion_voting}, and attention-based decision fusion \citep{C_2018ICML_attention_fusion}. 
Recently, uncertainty-based fusion strategies have gained popularity due to the reliability \citep{C_2020ICRA_fusion_UNO}. 

\begin{figure*}[t]
    \centering  
    \includegraphics[width=0.95\textwidth]{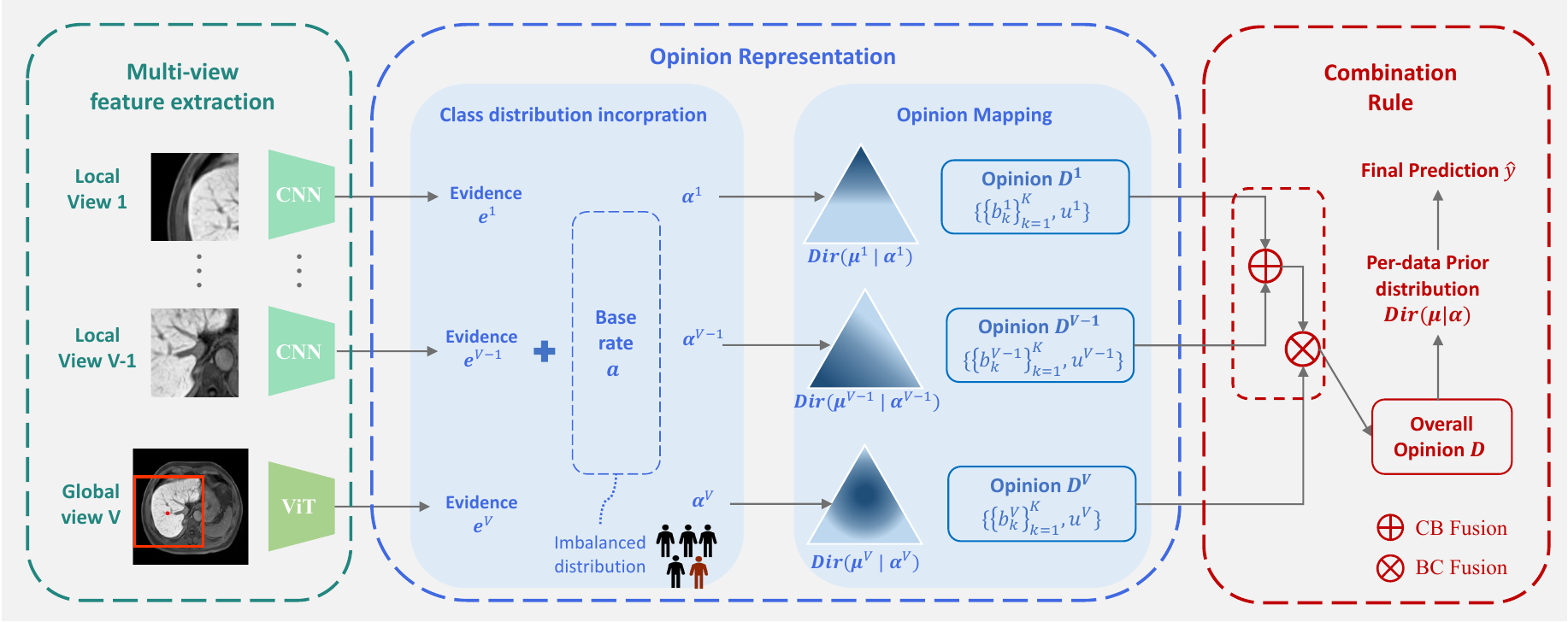}
    \caption{{The framework of MERIT consisted of multi-view feature extraction, opinion representation, and combination rule. In multi-view feature extraction, the whole liver MRI is converted to multiple local views and a global view image, which are encoded as evidence vectors $\{\bm{e}^v\}_{v=1}^V$ by convolutional networks (CNN) and vision transformer (ViT), respectively. In opinion representation, each evidence $\bm{e}^v$ combines with the class-distribution aware base rate $\bm{a}$ and generates a Dirichlet distribution $Dir(\bm{\mu}^v\mid\bm{\alpha}^v)$. They can be further represented by an opinion $\bm{D}^v$ composed of beliefs $b_k^v$ in each class and the uncertainty $u^v$. {In combination rule, the opinions are combined via  Cumulative Belief Fusion (CBF) and Belief Constraint Fusion (BCF) to derive the overall opinion $D$}, which could be converted to the per-data prior distribution $Dir(\bm{\mu}\mid\bm{\alpha})$ to achieve the final prediction $\hat{y}$.}}
    \label{fig:framework}
\end{figure*}

\subsection{Uncertainty Quantification}
Predictions made without uncertainty quantification could be unreliable \citep{J_2021_uncertainty_review}.
However, most deep-learning models are deterministic, and the uncertainty about their prediction could not be obtained directly. 
To estimate uncertainties, various methods have been proposed to apply Bayesian Neural Networks (BNNs), which replace the deterministic model parameters with distributions.
Among BNN-based methods, MC dropout is an efficient approach for computing predictive uncertainty. It randomly drops out weights to acquire samples from parameter distribution \citep{C_2016ICML_dropout, J_2023MIA_dropout}. 
VI-based methods approximated the posterior distribution of model parameters using a parameterized variational distribution by minimizing the evidence lower bound \citep{C_2019CVPR_Uncertainty_VI, J_2023MIA_VI}. 
Instead of modeling the parameter distribution, ensemble methods estimated uncertainties induced by the diversity of predictions from multiple sub-networks \citep{C_2021CVPR_masksembles, C_2022MICCAI_ensemble}.
Recent methods have applied Dirichlet distribution as a prior distribution for classification tasks to quantify distributional uncertainty.
PriorNet \citep{C_2018NIPS_prior_net} proposed to infer the underlying Dirichlet distribution by measuring the distance between in-distribution and out-of-distribution data.
Evidential learning \citep{C_2018NIPS_Evidential} modeled the prior distribution based on the theory of subjective logic.
Recently, it has also been applied to quantify the uncertainty in medical image analysis for reliable classification \citep{J_2021MIA_evi_class, C_2023ISBI_evi_class} and segmentation \citep{C_2022MICCAI_evi_seg, J_2024PAMI_evi_seg}.

\subsection{Evidence Theory}
Dempster-Shafer's (DS) evidence theory \citep{J_1968_evidence_theory} was proposed as a generalization of the Bayesian theory to express beliefs of elements in a state space. 
Different from probabilities, the belief function abandoned the sum principle to involve second-order uncertainty when the evidence is insufficient to support the belief.
DS evidence theory has been widely applied in multi-view learning \citep{J_2017_DS_multiview,J_2022_trusted_mv} since beliefs from multiple sources could be combined based on Dempster's combination rule. 
It was then extended to the subjective opinion model by considering the base rate, which makes it possible to define the mapping between subjective opinions and Dirichlet distribution \citep{B_2016_subjective}.
Based on such mapping, \citet{C_2018NIPS_Evidential} first applied the subjective opinion model in deep learning models to quantify distributional uncertainty.
Besides, the subjective logic theory also defines a set of interpretable fusion operators to combine opinions in different situations \citep{C_2012ICIF_fusion}.

\begin{table}[t]\caption{\label{tab:notation}{References for the mathematical symbols}} 
\small
\centering
\begin{tabular}{|p{0.16\columnwidth}|p{0.71\columnwidth}|}
\hline
Notations & Notions \\
\hline
$v, V$ & index and total number of views\\
$k,K$ & index and total number of stages \\
$\bm{x}^v$ & the image of view $v$\\
$y,\hat{y}$ & ground truth and predicted stage \\
$\bm{\mu}$ &  the probability of each class \\
$\bm{\alpha}$ & Concentration parameters of Dirichlet distribution \\
$f_{\bm{\theta}}(\cdot)$ & multi-view evidential model \\
$\bm{e}^v$ & the evidence vector of view \\
$\bm{D}, b_k, u$ & opinion, belief, and uncertainty \\
$\bm{a}, W $ & base rate and its weight\\
$N_k,N$ & number of samples of class k and all classes \\
$S$ & Dirichlet strength  \\
\hline
\end{tabular}
\end{table}

\section{Method}\label{sec:method}
This work is aimed at building a distributional uncertainty-aware classification model for liver fibrosis staging, which takes multiple patches of the liver MRI as input and predicts the staging results with quantified uncertainty.
Based on the subjective logic theory, our method represents the predictions of each view as opinions and incorporates logic-based combination rules to enhance the interpretability.

\subsection{Problem setup and overview}
Given images $\{\bm{x}^v\}_{v=1}^V$ from $V$ different views of liver MRI and the corresponding label $y \in\{1,\ldots, K\}$, our method aims to derive a distributional uncertainty-aware classification model defined by
\begin{equation}
    p(y\mid\{\bm{x}^v\}_{v=1}^V) = \int p(y\mid\bm{\mu})p(\bm{\mu}\mid\{\bm{x}^v\}_{v=1}^V) \,d\bm{\mu},
\end{equation}
where $p(y\mid\bm{\mu})=Mult(y\mid\bm{\mu})$ is the multinomial distribution with parameters $\bm{\mu}\in [0,1]^K$ denoting the probability of each class.
The latter term $p(\bm{\mu}\mid\{\bm{x}^v\}_{v=1}^V)$ is called per-data prior distribution \citep{C_2018NIPS_prior_net} over class probabilities $\bm{\mu}$, which captures the distributional uncertainty.
It is parameterized as follows,
\begin{equation}\label{eq:pdist}
    p(\bm{\mu}\mid\{\bm{x}^v\}_{v=1}^V)=Dir(\bm{\mu}\mid\bm{\alpha}),\quad \bm{\alpha}=f_{\bm{\theta}}(\{\bm{x}^v\}_{v=1}^V),
\end{equation}
where $Dir(\bm{\mu}\mid\bm{\alpha})= \frac{1}{B(\bm{\alpha})} \prod_{k=1}^K \mu_k^{{\alpha}_k}$ is the Dirichlet distribution and $B(\bm{\alpha})$ is the $K$-dimensional multinomial beta function.
$\bm{\alpha}\in \mathbb{R}^K_+$ are concentration parameters estimated by the multi-view evidential model $f_{\bm{\theta}}(\cdot)$ parameterized with $\bm{\theta}$.

As shown in Fig. \ref{fig:framework}, the multi-view evidential model $f_{\bm{\theta}}(\cdot)$ mainly consists of three parts, \textit{i.e.}, multi-view feature extraction, opinion representation, and combination rule. 
Within the feature extraction module, the local views (patches) and the global view (region of interest, ROI) of the liver image are encoded as evidence vectors $\{\bm{e}^v\}_{v=1}^V$ by evidential networks implemented with CNN and vision transformer (ViT), respectively. 
{Then, each evidence vector $\bm{e}^v$ is combined with the base rate $\bm{a}$ to form the Dirichlet distribution $Dir(\bm{\mu}^v\mid\bm{\alpha}^v)$, which can be mapped to the opinion $\bm{D}^v=\{\{b_k^v\}_{k=1}^K,u^v\}$ consisted of beleifs $b_k^v$ in each class and the uncertainty $u^v$.}
{In the final stage, interpretable fusion operators are applied to combine opinions from all views, \textit{i.e.}, $\{\bm{D}^v\}_{v=1}^{V}$, and generate the overall opinion, which can be transformed into the per-data distribution $Dir(\bm{\mu}\mid\bm{\alpha})$ defined in Eq. (\ref{eq:pdist}), and yield the final prediction by $\hat{y}=\text{arg max}_k \mathbb{E}_{Dir(\bm{\mu}|\bm{\alpha})}[\mu_k]$.}
{The mathematical notations are summarized in Table \ref{tab:notation}.} 
The details of opinion representation, combination rule, multi-view feature extraction, and the training paradigm will be discussed in Sec.~\ref{sec:opinion}, Sec.~\ref{sec:combine}, Sec.~\ref{sec:mv}, and Sec.~\ref{sec:train}, respectively.

\subsection{Opinion representation based on subjective logic}\label{sec:opinion}
To enhance the reliability in both feature and class distribution shift scenarios, we propose to quantify the distributional uncertainty in $Dir(\bm{\mu}^v\mid\bm{\alpha^v})$ based on evidence theories \citep{C_2018NIPS_Evidential}.

{In the evidence theory, the evidence vector $\bm{e}^v\in\mathbb{R}_+^K$ is used in replacement of probabilities $\bm{\mu}^v$ to express beliefs of possible states (\textit{i.e.,} fibrosis stages). Each evidence vector could be mapped to a Dirichlet distribution $Dir(\bm{\mu}^v\mid\bm{\alpha^v})$, namely the distribution of class probabilities $\bm{\mu}^v$, given the base rate $\bm{a}$ which represents prior knowledge (\textit{e.g.}, previous class distribution). Hence, the evidence vector could reflect the uncertainty of the predicted class probabilities $\bm{\mu}^v$, denoted as the distributional uncertainty in this work.
The uncertainty could be quantified according to the subjective logic theory \citep{B_2016_subjective}, in which $Dir(\bm{\mu}^v\mid\bm{\alpha^v})$ is mapped to the opinion $\bm{D}^v=\{\{b_k^v\}_{k=1}^K,u^v\}$ where $\{b_k^v\}_{k=1}^K$ express beliefs in each class and $u^v$ is the quantified uncertainty.
The details of the mapping between evidence, Dirichlet distribution, and opinions are elaborated in the following sections.}

\subsubsection{Evidence representation of Dirichlet distribution}
Given the evidence $\bm{e}^v\in\mathbb{R}_+^K$ and base rate $\bm{a}=[a_1,\cdots,a_K]\in[0,1]^K$, the concentration parameter $\bm{\alpha}^v$ of the Dirichlet distribution in view $v$ could be expressed as,
\begin{equation}\label{eq:evidence}
\bm{\alpha}^v = \bm{e}^v + \bm{a}W,
\end{equation}
where the base rate satisfies $\sum_{k=1}^K a_k=1$, and $W>0$ is the hyper-parameter used to balance $\bm{e}^v$ and $\bm{a}$, typically set equal to the number of classes $K$. 
In this work, we assume that the base rate remains the same across all views, calculated by,
\begin{equation}\label{eq:prior}
    a_k = \frac{N_k}{N}, \quad \forall k\in\{1,\ldots,K\},
\end{equation}
where $N_k$ and $N$ are the number of samples for class $k$ and all classes, respectively.
According to Eq. (\ref{eq:evidence}) and Eq. (\ref{eq:prior}), the concentration parameter $\bm{\alpha}^v$ is determined by both the evidence estimated from the image and the class distribution.

\begin{figure}[t]
    \centering 
    \includegraphics[width=0.95\columnwidth]{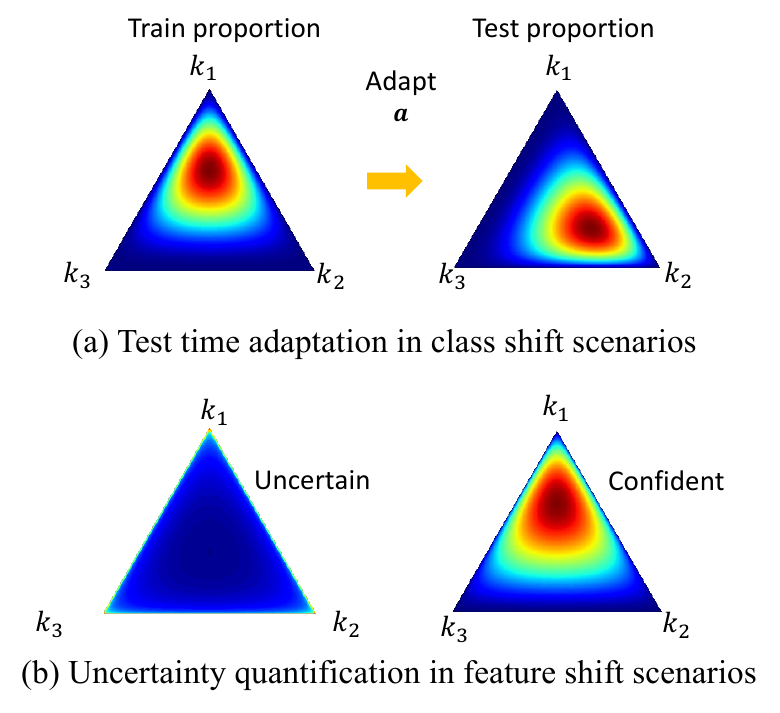} 
    \caption{The predicted Dirichlet distribution under different distribution shift scenarios. (a) The predicted Dirichlet distribution can be adapted by modifying the base rate with the estimated test class proportion. (b) Out-of-distribution sample (left), which could not provide any evidence for the decision, would result in a uniform Dirichlet distribution and yield high uncertainty in our model, while the sample without feature shift (right) would produce low uncertainty.}
\label{fig:dist_shift}
\end{figure}

There are two advantages to employing the class distribution-aware representation of $\bm{\alpha}^v$.
Firstly, when the class proportion of the training set is imbalanced, it enables the evidential network to allocate equal attention to the feature representation of all classes during training.
In contrast, using a uniform base rate, as done in previous work \cite{C_2018NIPS_Evidential}, would result in the network focusing more on the majority class. 
Secondly, it allows for test-time adaptation, which could enhance performance in scenarios involving class distribution shifts. 
As shown in Fig. \ref{fig:dist_shift}(a), the predicted Dirichlet distribution can be adjusted by modifying the base rate from the class proportion of the training data to an estimated proportion of the test data.

\subsubsection{Mapping between Dirichlet distribution and opinion}
According to the theory of subjective logic, each Dirichlet distribution $Dir(\bm{\mu}^v\mid{\bm{\alpha}^v)}$ is associated with an opinion $\bm{D}^v=\{\{b_k^v\}_{k=1}^K,u^v\}$ satisfying the equation,
\begin{equation}
    u^v+\sum_{k=1}^K{b_k^v}=1,
\end{equation}
where $b_k^v$ is the belief mass for class $k$ and $u^v$ measures the uncertainty of the Dirichlet distribution.
The opinion could be derived from the Dirichlet distribution by,
\begin{equation}\label{eq:opinion}
    b_k^v=\frac{\alpha_k^v-{a_k}W}{S^v}, \quad u^v=\frac{W}{S^v},
\end{equation}
where $S^v=\sum_{k=1}^K{\alpha_k^v}$ is the Dirichlet strength. The derivation of Eq. (\ref{eq:opinion}) is based on the definition of projection from the opinion space to the probability space, \textit{i.e.}, $\mathbb{E}[Dir(\bm{\mu}^v\mid{\bm{\alpha}^v)}]=\bm{b}^v+\bm{a}u^v$. Please refer to the supplementary materials for more details.

Fig. \ref{fig:dist_shift}(b) demonstrates the rationale behind the uncertainty quantification in our method.
We assume that samples with extreme feature shifts can not provide any evidence for decision-making, \textit{i.e.,} $\bm{e}^v=\bm{0}$, and result in a uniform Dirichlet distribution.  
According to Eq. (\ref{eq:evidence}) and Eq. (\ref{eq:opinion}), the uncertainty $u^v$ has a negative correlation with the sum of evidence. Therefore, such samples would yield high uncertainty in our model.

\subsection{Interpretable combination rule based on DS evidence theory} \label{sec:combine}
To ensure interpretability in the decision-making process based on opinions, 
we use belief fusion operators to combine opinions derived from multiple views, following the DS evidence theory \citep{C_2012ICIF_fusion}. 

Given opinions from local views $\{\bm{D}^v\}_{v=1}^{V-1}$ and the global view $\bm{D}^V$, we derive the combined opinion by using the following combination rule,
\begin{equation}\label{eq:combine}
    \bm{D} = (\bm{D}^1\oplus\cdots\oplus \bm{D}^{V-1})\otimes \bm{D}^V,
\end{equation}
where $\oplus$ and $\otimes$ denote the reduced cumulative belief fusion (CBF) operator and belief constraint fusion (BCF) operator, respectively.
These two operators are defined as follows \citep{B_2016_subjective},

\begin{definition}[Cumulative Belief Fusion Operator]
The combined opinion $\bm{D}=\{\{b_k\}_{k=1}^K,u\}$, calculated from opinion $\bm{D}^m=\{\{b_k^m\}_{k=1}^K,u^m\}$ and opinion $\bm{D}^n=\{\{b_k^n\}_{k=1}^K,u^n\}$ using cumulative fusion operator (\textit{i.e.,} $\bm{D}=\bm{D}^m \oplus \bm{D}^n$), is derived as follows,
\begin{equation}\label{eq:CBF}
    b_k= \frac{b_k^m u^n + b_k^n u^m}{u^m+u^n-u^m u^n}, \quad u=\frac{u^m u^n}{u^m+u^n-u^m u^n},
\end{equation}
where the uncertainties are supposed to satisfy $u^m+u^n\neq 0$. 
\end{definition}
\begin{definition}[Belief Constraint Fusion Operator]
The combined opinion $\bm{D}=\{\{b_k\}_{k=1}^K,u\}$, calculated from opinion $\bm{D}^m=\{\{b_k^m\}_{k=1}^K,u^m\}$ and opinion $\bm{D}^n=\{\{b_k^n\}_{k=1}^K,u^n\}$ using belief constraint fusion operator (\textit{i.e.,} $\bm{D}=\bm{D}^m \otimes \bm{D}^n$), is derived as follows,
\begin{equation}\label{eq:BCF}
    b_k= \frac{1}{C}(b^m_k b^n_k + b_k^m u^n + b_k^n u^m), \quad u=\frac{1}{C}u^m u^n,
\end{equation}
where $C=1-\sum_{i\neq j}{b_i^m b_j^n}$ is the normalization factor satisfying $C\neq 0$. 
\end{definition}

The combined opinion $\bm{D}$ that expresses the overall belief and uncertainty could be converted to the final Dirichlet distribution $Dir(\bm{\mu}\mid\bm{\alpha})$ through the relationship in Eq. (\ref{eq:opinion}).

\subsubsection{Interpretation of the combination rule} 
In Eq. (\ref{eq:combine}), the opinions from local views and the global view are combined using distinct fusion operators, according to the sources of opinions and the purpose of the fusion. 

For the combination of local views, we assume each view contains incomplete information for decision making, and combining two views will result in the decrease of uncertainty (\textit{i.e.}, an increase in evidence).
This assumption aligns with the scenario where the CBF operator applies, as when all views share the same base rate, CBF is equivalent to adding up the evidence.

Formally, the evidence $\bm{e}$ associated with the combined opinion $\bm{D}=\bm{D}^m \oplus \bm{D}^n$ can be written as,
\begin{equation}\label{eq:add_evidence}
    \bm{e} = \bm{e}^m + \bm{e}^n,
\end{equation}
where $\bm{e}^m$ and $\bm{e}^n$ are the evidence vectors associated with $\bm{D}^m$ and $\bm{D}^n$, respectively. Eq. (\ref{eq:add_evidence}) can be derived by combining Eq. (\ref{eq:evidence}), Eq. (\ref{eq:opinion}) and Eq. (\ref{eq:CBF}). Please refer to the supplementary materials for more details.

When combining the opinion of the global view with the fused opinion of local views, 
we assume the opinions are made with different preferences based on the complete information from the entire liver image, due to distinct feature extraction networks.
Therefore, the purpose of the fusion is to derive a new opinion that is acceptable for both views.
The BCF operator could produce the most acceptable results due to the following properties, 
\begin{itemize}
    \item If both opinions are equally confident, the combined opinion believes in the class that both opinions agree on.
    \item If one opinion is uncertain, the belief of the combined opinion is more similar to the confident one.
    \item If both opinions are uncertain, the combined opinion is uncertain.
\end{itemize}

Theoretically, let $\hat{k}$ denote the class that the combined opinion $\bm{D}=\bm{D}^m \otimes \bm{D}^n$ believes in, \textit{i.e.,} $\hat{k}=\text{arg max}_j b_j$, and the class that both opinions agree on is defined as $\Tilde{k}=\text{arg max}_j b^m_j b^n_j$.
The above properties are supported by the following propositions,
\begin{proposition}
    If both opinions are equally confident, \textit{i.e.,} $u^m=u^n< \dfrac{b^m_{\Tilde{k}}b^n_{\Tilde{k}}- b^m_j b^n_j}{\vert b^m_j+b^n_j-(b^m_{\Tilde{k}}+b^n_{\Tilde{k}})\vert}$ for $j\neq \Tilde{k}$, the combined opinion believes in the class that both opinions agree on, \textit{i.e.,} $\hat{k}=\Tilde{k}$.
\end{proposition}
\begin{proposition}
 The dissimilarity between the beliefs of $\bm{D}^m$ and the combined opinion $\bm{D}$, \textit{i.e.,}  $b^m_{j} - b_{j}$, is negatively correlated with the uncertainty of opinion $\bm{D}^n$, 
 $\forall j=1,\ldots,K$ that satisfies $ b^m_{j} - b_{j}>0$.
 As $u^n$ approaches $1$, $b^m_{j} - b_{j}$ decreases to $0$.
\end{proposition}

For the last property, it can be observed from Eq. (\ref{eq:BCF}) that $u$ positively correlated with $u^m u^n$, which means that the uncertainty of the combined opinion is large when both opinions are uncertain. For better comprehension, we provide several examples in the supplementary material to demonstrate the properties above.

\begin{figure}[t]
    \centering
    \begin{tabular}{c}
       \includegraphics[width=0.9\columnwidth]{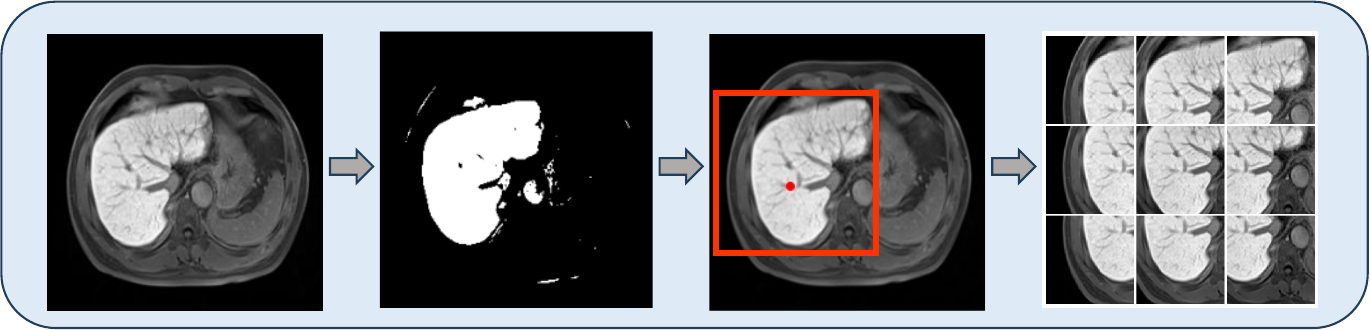}  \\
       (a) Multi-view extraction \\
        \includegraphics[width=0.9\columnwidth]{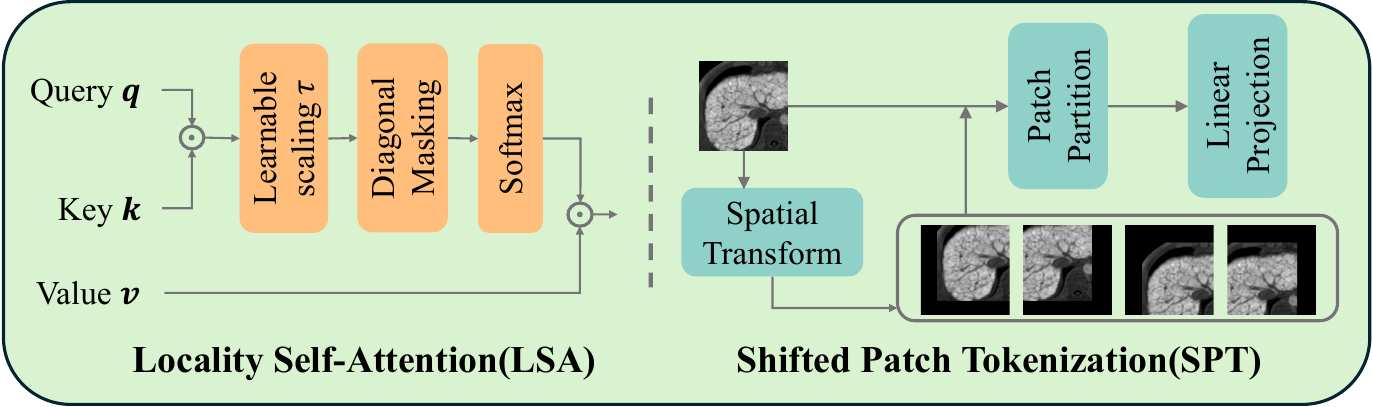}\\
        (b) Data-efficient transformer 
    \end{tabular}
    \caption{{(a) The pipeline to extract sub-views of the liver. First, the foreground is extracted using intensity-based segmentation. Based on the segmentation, a square ROI centered at the centroid of the liver is cropped. Then overlapped sliding windows are used in the ROI to obtain nine sub-views of the liver. (b) Locality self-attention (LSA) and shifted patch tokenization (SPT) modules applied in the data-efficient transformer. SPT replaces the original self-attention in ViT, which introduces a learnable parameter $\tau$ to scale the unnormalized attention map (\textit{i.e.,} $(\bm{qk}^T)/\tau$), whose diagonal elements are then replaced with constants before softmax.
    LSA modifies the tokenization strategy of ViT by concatenating spatially transformed images with the original one before patch partition.}}
    \label{fig:multiview}
\end{figure}

\subsection{Modeling multi-view features}  \label{sec:mv}

To exploit informative features across the whole liver, we extract image patches and consider each patch as a local view. The pipeline for view extraction is shown in Fig. \ref{fig:multiview}(a). A square ROI is cropped based on the segmentation of the foreground. Then nine sub-views of the liver are extracted in the ROI through overlapped sliding windows.
Furthermore, the whole ROI is incorporated as the global view to assist in the extraction of features.

\subsubsection{Feature extraction from multi-views} 
For local views, we choose a CNN architecture to learn representations that focus on detailed features, employing Softplus as the activation to ensure non-negative evidence output.  

To capture the global representation, we utilize a data-efficient transformer for the global view. 
We follow \cite{J_2022_VIT_small} and enhance the efficacy of the transformer on small datasets by increasing locality inductive bias, \textit{i.e.}, the assumption about relations between adjacent pixels.
The conventional ViT \citep{C_2021ICLR_vit}, without such assumptions, demands more training data compared to convolutional networks \citep{C_2020NIPS_inductive_bias}. 
Therefore, we adopt the shifted patch tokenization (SPT) and locality self-attention (LSA) strategy to improve the locality inductive bias.

As shown in Fig. \ref{fig:multiview}(b), SPT diverges from the standard tokenization by offsetting the input image in four diagonal directions by half of the patch size. These shifted images are then concatenated with the original images along the channel dimension, enriching the exploitation of spatial correlation among neighboring pixels. 
Subsequently, the concatenated images are partitioned into patches and linearly projected as visual tokens, similar to the process employed by ViT.

LSA enhances the self-attention mechanism in ViT by sharpening the distribution of the attention map to prioritize important visual tokens. As shown in Fig. \ref{fig:multiview}(b), diagonal masking and temperature scaling are performed before applying softmax to the attention map. Given the input feature $\bm{X}$, The LSA module is formalized as,
\begin{equation}
    L(\bm{X})= \text{softmax}(\mathcal{M}(\bm{qk}^T)/\tau)\bm{v},
\end{equation}
where $\bm{q,k,v}$ are the query, key, and value vectors, respectively, derived from linear projections of $\bm{X}$. $\mathcal{M}$ denotes the diagonal masking operator that sets the diagonal elements of $\bm{qk}^T$ to a small number (\textit{e.g.},$-\infty$). $\tau\in \mathbb{R}$ is a learnable scaling factor.

\subsection{Training paradigm}\label{sec:train}

Theoretically, the proposed MERIT can be trained in an end-to-end manner. 
For each view $v$, we use the integrated cross-entropy loss as in \citet{C_2018NIPS_Evidential},
\begin{equation}
    \begin{aligned}
        \mathcal{L}_{ice}^{v}
        &=\mathbb{E}_{\bm{\mu}^{v}\sim Dir(\bm{\mu}^{v}|\bm{\alpha}^{v})}[\mathcal{L}_{CE}(\bm{\mu}^{v},\bm{y})]
        \\
        &=\int\left[\sum_{k=1}^K-y_{k}\log(\mu_{k}^v)\right]\frac{1}{B(\boldsymbol{\alpha}^{v})}\prod_{k=1}^K (\mu_{k}^v)^{\alpha_{k}^v-1}\mathrm{d}\bm{\mu}^v
        \\
        &=\sum_{k=1}^Ky_{k}(\psi(S^{v})-\psi(\alpha_{k}^{v})),
    \end{aligned}
    \label{eq:ice}
\end{equation}
where $\psi$ is the digamma function and $\bm{y}$ is the one-hot label. 

The loss function in Eq. (\ref{eq:ice}) deals with distribution shifts in two aspects. 
For class shift, the class with fewer samples would induce a lower base rate $a_k$, resulting in a lower $\alpha_k^v$ according to Eq. (\ref{eq:evidence}) and Eq. (\ref{eq:prior}). As $\psi(\cdot)$ is monotonically increasing, the class distribution-aware base rate will incur a greater penalty, which will increase the likelihood that the model correctly classifies samples from the minority class \citep{J_2019_class_imbalance_survey}. 
For samples with feature shift, the network tends to yield a smaller amount of evidence, \textit{i.e.}, $S^v$, instead of the increase of corresponding concentration parameter $\alpha_k^v$, to minimize the overall loss \citep{C_2018NIPS_Evidential}. Therefore, the loss function will increase the uncertainty for such samples, according to Eq. (\ref{eq:opinion}). 

We also apply a regularization term to increase the uncertainty of misclassified samples,
\begin{equation}
    \begin{aligned}
        KL[Dir(\bm{\mu}^{v}\mid\tilde{\bm{\alpha}}^{v})||Dir(\bm{\mu}^{v}\mid\bm{\beta})]
        & = \log\left(\frac{\Gamma\Bigl(\sum_{k=1}^{K}\tilde{\alpha}_{k}^{v}\Bigr)\prod_{k=1}^{K}\Gamma({\beta}_{k})}{\Gamma\Bigl(\sum_{k=1}^{K}{\beta}_{k}\Bigr)\prod_{k=1}^{K}\Gamma(\tilde{\alpha}_{k}^{v})}\right)  
        \\
         +\sum_{k=1}^K&\left(\tilde{\alpha}_{k}^{v}-\beta_{k}\right)\left[\psi\left(\tilde{\alpha}_{k}^{v}\right)-\psi\left(\sum_{k=1}^K\tilde{\alpha}_{k}^{v}\right)\right],
    \end{aligned}
\end{equation}
{where $\bm{\tilde{\alpha}}^{v}= \bm{y} \odot \bm{\beta}+(\bm{1}-\bm{y})\odot \bm{\alpha}^v$, avoiding unnecessary penalty for correct-classified samples and $\Gamma(\cdot)$ denotes the gamma function}. $Dir(\bm{\mu}^v\mid\bm{\beta})$ is the non-evidence prior distribution, \textit{i.e.}, $\bm{\beta}=\bm{a}\cdot W$. 
For each $v$, the total loss is,
\begin{equation}
    \mathcal{L}^{v}=\mathcal{L}_{ice}^{v}+\lambda KL[Dir(\bm{\mu}^{v}\mid\tilde{\bm{\alpha}}^{v})||Dir(\bm{\mu}^{v}\mid\bm{\beta})],
\end{equation}
where $\lambda$ is the balance factor that gradually increases during training. 
The overall loss is the summation of losses from all views and the loss for the combined opinion,
\begin{equation}\label{loss}
     \mathcal{L}_{Overall} = \mathcal{L}_{Combined} + \sum_{k=1}^{V}\mathcal{L}^{v},
\end{equation}
where the loss for the combined opinion $\mathcal{L}_{Combined}$ is  implemented in the same way as $\mathcal{L}^{v}$.

\begin{table*}[t]
    \centering
    \caption{Comparison with uncertainty-aware methods. The expected calibration error (ECE) was evaluated in addition to ACC and AUC. Methods with lower ECE are more reliable.}
    \resizebox{1\textwidth}{!}{
        \begin{tabular}{ccccccc}
        \hline
        \multirow{2}{*}{Method} &\multicolumn{3}{c}{Cirrhosis(S4 vs S1-3)} & \multicolumn{3}{c}{Significant Fibrosis(S1 vs S2-4)}\\
        \cline{2-7}
        \multirow{2}{*}{}                        &ACC $\uparrow$  & AUC $\uparrow$ & ECE $\downarrow$ &ACC $\uparrow$&AUC $\uparrow$&ECE $\downarrow$\\
        \hline
    {Softmax} \citep{A_2021_Softmax}                 & $74.4\pm1.95$ & $75.7\pm1.63$ & $0.238\pm0.015$ & $74.7\pm2.41$ & $76.2\pm5.02$  & $0.261\pm0.024$ \\
    \hdashline[8pt/5pt]
    {Ensemble} \citep{C_2021CVPR_masksembles}       & $72.0\pm2.67$ & $75.8\pm2.65$ & $0.200\pm0.045$ & $72.0\pm4.45$ & $74.8\pm4.71$ & $0.169\pm0.047$  \\
    {Dropout} \citep{C_2016ICML_dropout}            & $70.0\pm2.51$ & $76.2\pm3.67$ & $0.265\pm0.027$ & $71.0\pm1.96$ & $75.8\pm3.62$ & $0.263\pm0.026$  \\
    {VI} \citep{C_2019CVPR_Uncertainty_VI}          & $69.1\pm3.08$ & $75.3\pm3.19$ & \bm{$0.050\pm0.037$} & $64.5\pm2.76$ & $69.3\pm5.73$ & \bm{$0.077\pm0.019$}  \\
    {PriorNet} \citep{C_2018NIPS_prior_net}      & $67.9\pm2.84$ & $73.7\pm3.31$ & $0.218\pm0.022$ & $64.4\pm3.32$ & $70.6\pm1.43$   & $0.212\pm0.033$  \\
        \hdashline[8pt/5pt]
    {MERIT(w/o)~\citep{C_2023MICCAI_ours}}  & {{$79.6 \pm 3.39$}}  & {$82.3\pm1.69$} & {$0.159\pm0.050$}  & {{$82.9\pm1.91$}} & {$82.4\pm2.01$}  &  {$0.145\pm0.035$} \\
    MERIT (Ours)     & \bm{$80.4\pm1.53$} & \bm{$85.5\pm1.54$} & {$0.100\pm0.012$} & \bm{$84.1\pm2.84$} & \bm{$87.4\pm3.27$} & {$0.154\pm0.029$}  \\        \hline
        \end{tabular}
        }
        \label{tab:uncertainty}
\end{table*}

\section{Experiment}\label{sec:experiment}

In this section, we evaluated the effectiveness of MERIT on multi-center liver MRI datasets. 
To demonstrate the reliability, we first compared the calibration performance of MERIT with other uncertainty-aware methods in Sec. \ref{sec:cali}. 
Then we validated the applicability in two distribution shift scenarios in Sec. \ref{sec:feature_shift} and Sec. \ref{sec:label_shift}. 
For the study of interpretability, we first performed comparison experiments on the interpretable combination rules in Sec. \ref{sec:rule}. 
The post hoc interpretation provided by MERIT was analyzed in Sec. \ref{sec:post}. 
Finally, comparisons with other multi-view learning methods were presented in Sec. \ref{sec:multiview}.

\subsection{Dataset}
The proposed MERIT was evaluated on a public multi-center Gd-EOB-DTPA-enhanced hepatobiliary phase MRI dataset 
including $703$ patients acquired from four scanners, \textit{i.e.}, Philips 1.5T, Philips 3.0T, Siemens 1.5T and Siemens 3.0T. The gold standard was obtained through the pathological analysis of the liver biopsy or liver resection within three months before and after MRI scans. 
A total of $155$ patients were identified with fibrosis stage S1, $91$ with S2, $76$ with S3, and $381$ with the most advanced stage, S4. 
Following \citet{J_2018R_pmx_liver}, the slices with the largest liver area in images were selected. The data were then preprocessed with z-score normalization, resampled to a resolution of $1.5 mm\times 1.5 mm$, and cropped to $256 \times 256$ pixel. For multi-view extraction, the size of the ROI, window, and stride were $160$, $96$, and $32$, respectively. A four-fold cross-validation strategy was employed in the comparison experiments. 

Results of two tasks with clinical significance \citep{J_2018R_pmx_liver} were evaluated, \textit{i.e.}, staging cirrhosis (S4 vs S1-3) and identifying substantial fibrosis (S1 vs S2-4). 
Notably, in the task of identifying substantial fibrosis, the classes were imbalanced. 
MERIT utilized a class distribution-aware base rate. 
For fairness in the comparison study, we implemented over-sampling of the S1 data and under-sampling of the S4 data during the experiments of staging substantial fibrosis.

\begin{figure}[t]
    \centering
    \includegraphics[width=0.47\textwidth]{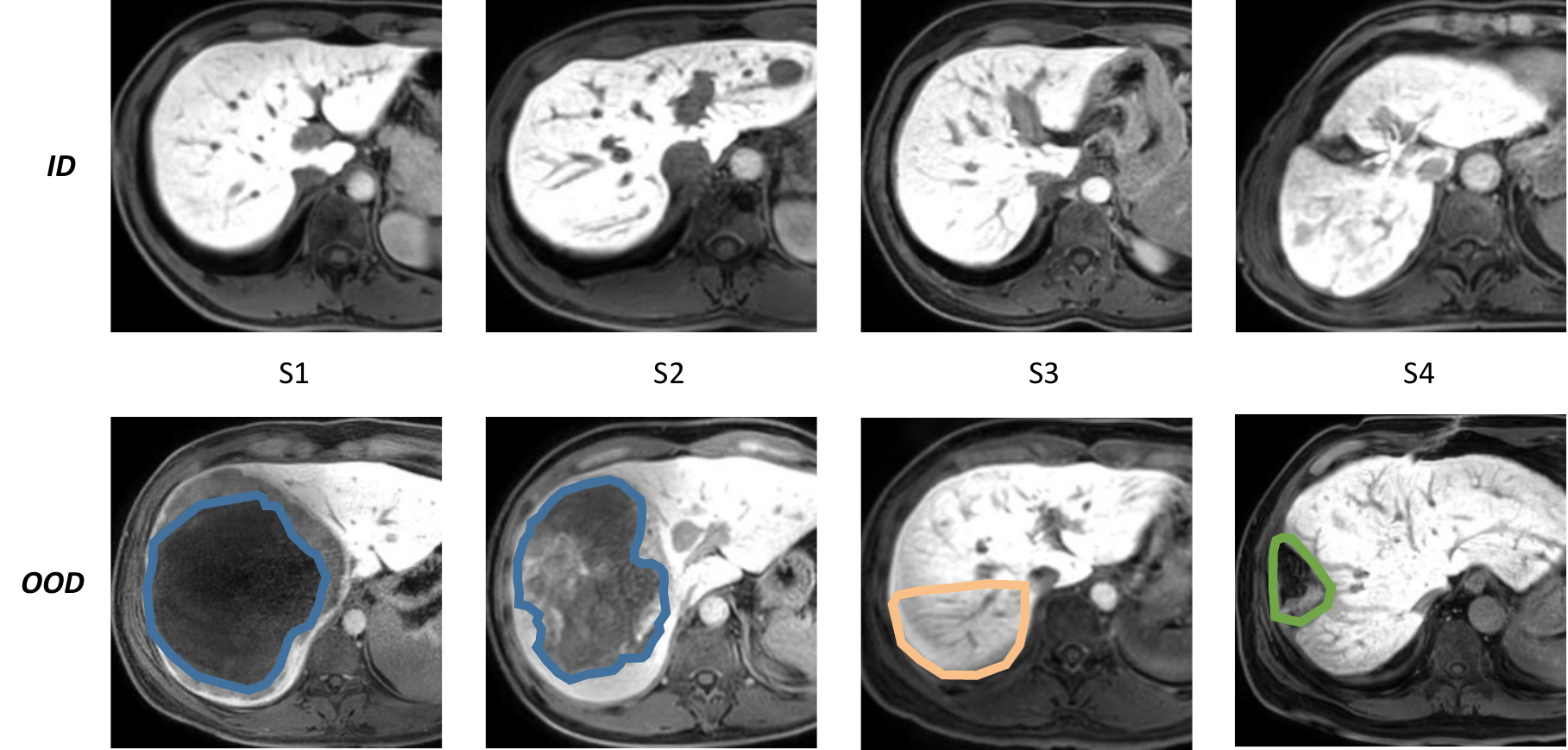}
    \caption{MRI scans of liver tissue illustrating ID and OOD data. The top row (ID) displays liver slices with an effective area greater than $90\%$ denoting scans considered as normal for the training dataset. The bottom row (OOD) depicts liver slices with an effective area less than $90\%$, representing anomalous cases used for testing the out-of-distribution dataset. Specifically, the blue area indicates liver mass, the orange area indicates local artifact and the green area indicates previous liver surgery.}
    \label{fig:OOD_data}
\end{figure}

\subsection{Implementation details}
Augmentations such as random rescale, flip, and cutout \citep{A_2017_cutout} were applied during training. We chose ResNet34 \citep{C_2016CVPR_resnet} for feature extraction of local views. The framework was trained using Adam optimizer with an initial learning rate of $1e-4$ for $500$ epochs, which was decreased by using the polynomial scheduler. The balance factor $\lambda$ was set to increase linearly from 0 to 1 during training. 
For local views, we used the model weights pre-trained on ImageNet, and the transformer was pre-trained on the global view images for $200$ epochs using the same setting. The framework was implemented using Pytorch and was run on one Nvidia RTX 3090 GPU.

\begin{table*} [t]
    \centering
    \caption{Comparison with uncertainty-aware methods, which highlights MERIT's superior performance in both OOD detection and staging tasks, demonstrating our advanced capabilities in quantifying distributional uncertainty and achieving exceptional generalization.}
    \resizebox{1\textwidth}{!}{
        \begin{tabular}{cc:ccccccc}
        \hline
        \multirow{2}{*}{Method} &\multicolumn{1}{c:}{OOD Detection} & \multicolumn{3}{c}{ID Staging} & \multicolumn{3}{c}{OOD Staging}\\
        \cline{2-8}
        \multirow{2}{*}{}      & ACC $\uparrow$ & ACC $\uparrow$ &  AUC$\uparrow$ & ECE $\downarrow$& ACC $\uparrow$&AUC $\uparrow$ & ECE$\downarrow$\\
        \hline
    {Softmax} \citep{A_2021_Softmax}              & $35.3\pm4.40$ & $64.8\pm3.42$ & $69.3\pm2.52$ & $0.333\pm0.029$ & $59.1\pm2.60$ & $59.7\pm1.85$ & $0.390\pm0.016$ \\
    \hdashline[8pt/5pt]
    {Ensemble} \citep{C_2021CVPR_masksembles}    & $39.3\pm4.54$ &  $64.8\pm2.79$ & $71.1\pm2.26$ & $0.320\pm0.019$ & $57.3\pm3.21$ & $55.9\pm4.37$ & $0.390\pm0.025$   \\
    {Dropout} \citep{C_2016ICML_dropout}         & $42.2\pm5.96$ & $67.5\pm2.06$ & $72.9\pm1.86$ & $0.300\pm0.016$ & $57.7\pm2.34$ & $56.9\pm2.93$ & $0.390\pm0.027$    \\
    {VI} \citep{C_2019CVPR_Uncertainty_VI}      & $51.8\pm2.27$ & $69.8\pm2.28$  &$76.4\pm1.76$ & \bm{$0.057\pm 0.024$} & $57.5\pm3.10$ & $58.2\pm2.96$ & \bm{$0.133\pm0.028$}   \\
    {PriorNet} \citep{C_2018NIPS_prior_net}      & $56.7\pm3.62$ & $65.5\pm3.20$  &$73.8\pm2.79$ & $0.193\pm 0.029$ & $54.2\pm2.05$ & $56.9\pm2.08$ & $0.305\pm0.022$   \\
        \hdashline[8pt/5pt]
    MERIT (Ours)                                           & $\bm{70.6\pm3.63}$ & $\bm{81.3\pm2.62}$ & $\bm{87.4\pm0.81}$ & $0.120\pm0.004$  & $\bm{70.1\pm1.49}$ & $\bm{74.1\pm0.89}$ &$0.166\pm0.014$\\
        \hline
        \end{tabular}
        }
        \label{tab:OOD_detection}
\end{table*}

\begin{figure*}
    \centering
    \begin{tabular}{ccc}
      \includegraphics[width=0.32\textwidth]{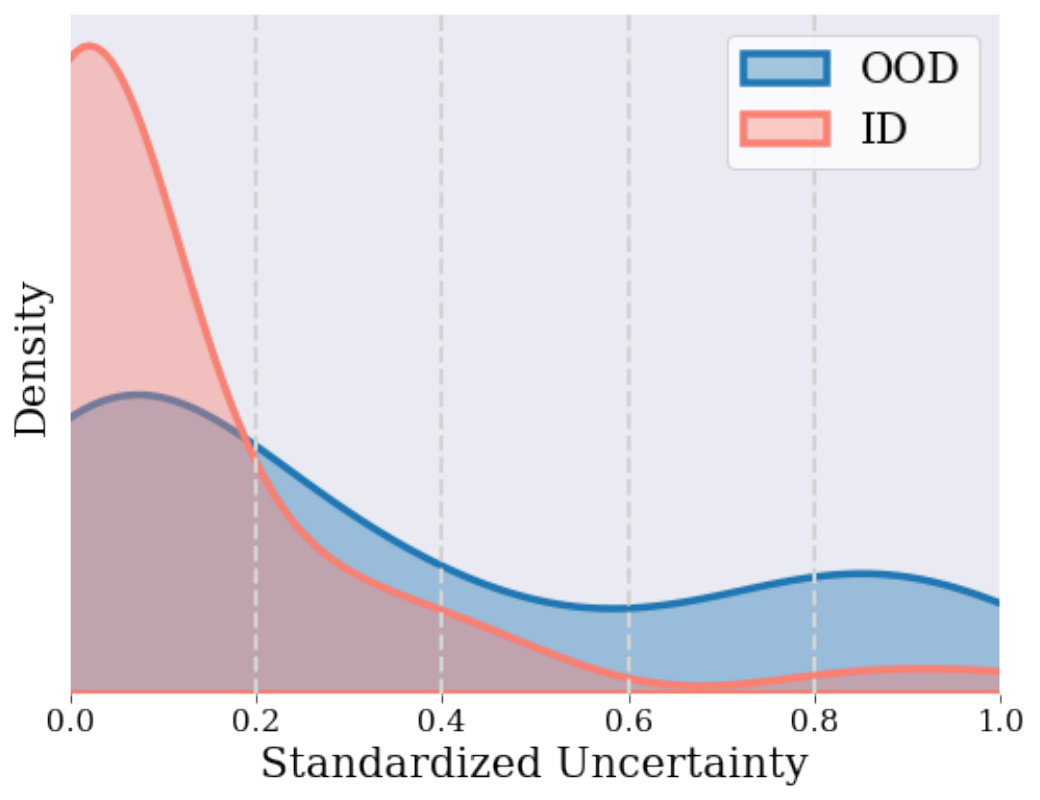}
      &
      \includegraphics[width=0.32\textwidth]{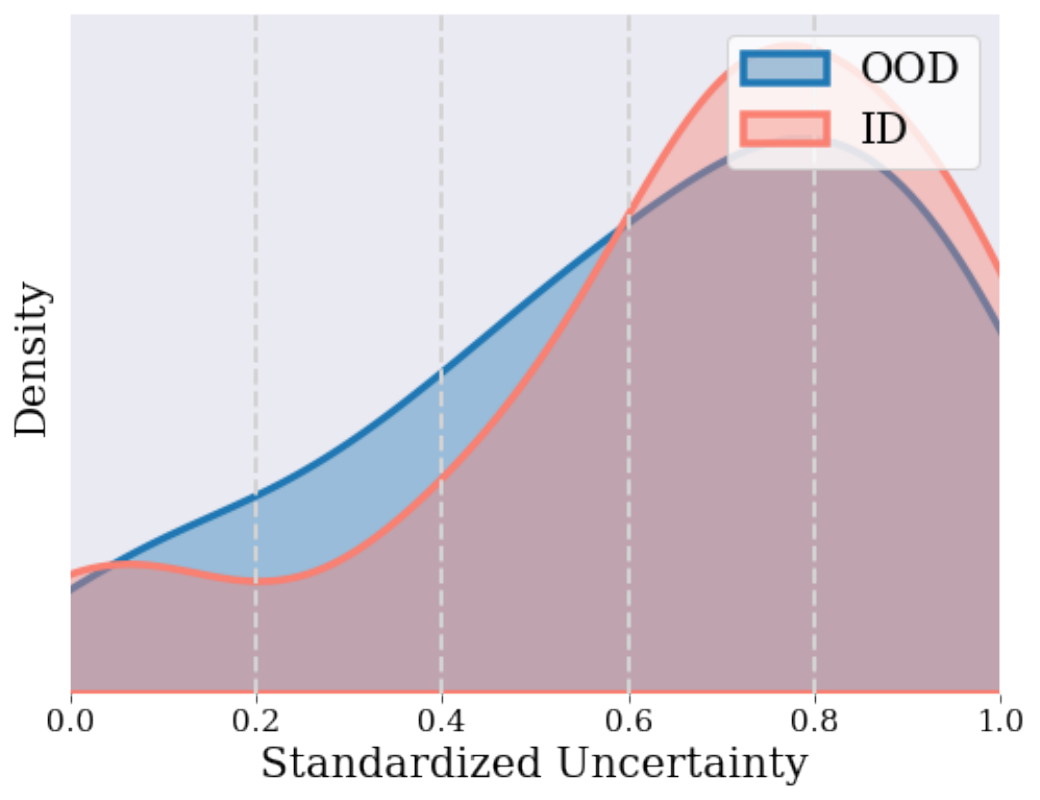}
      &
      \includegraphics[width=0.32\textwidth]{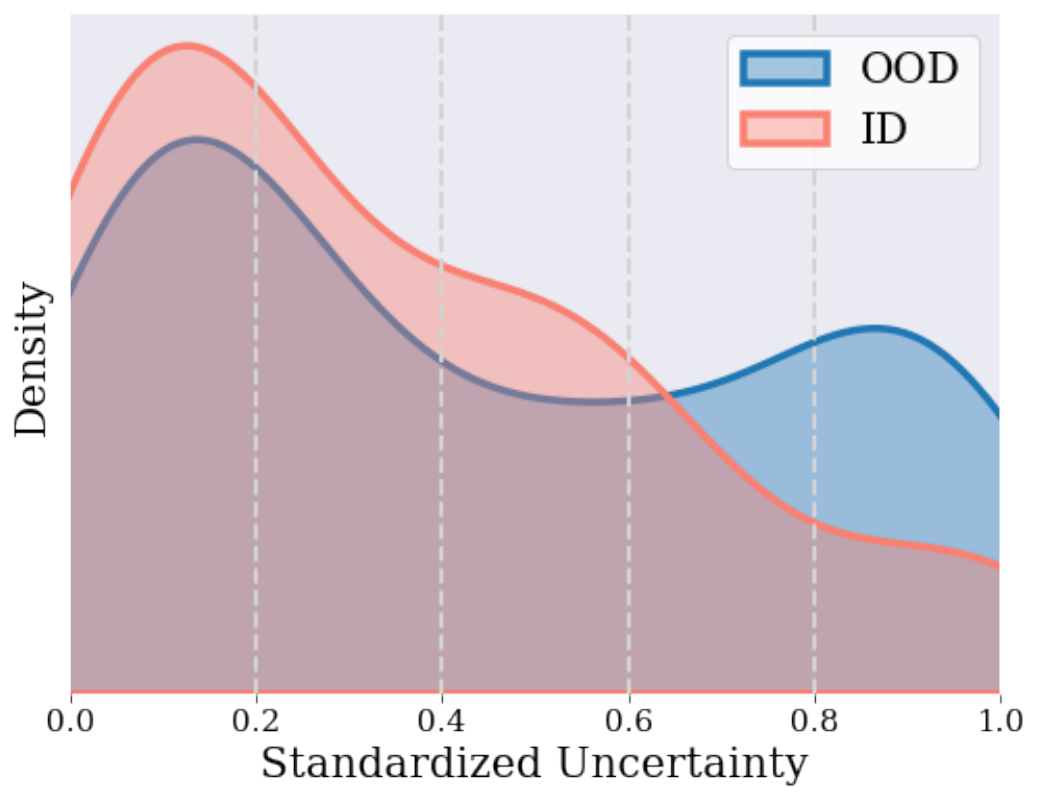}
      \\
      (a) Softmax & (b) Ensemble & (c) Dropout\\
      \includegraphics[width=0.32\textwidth]{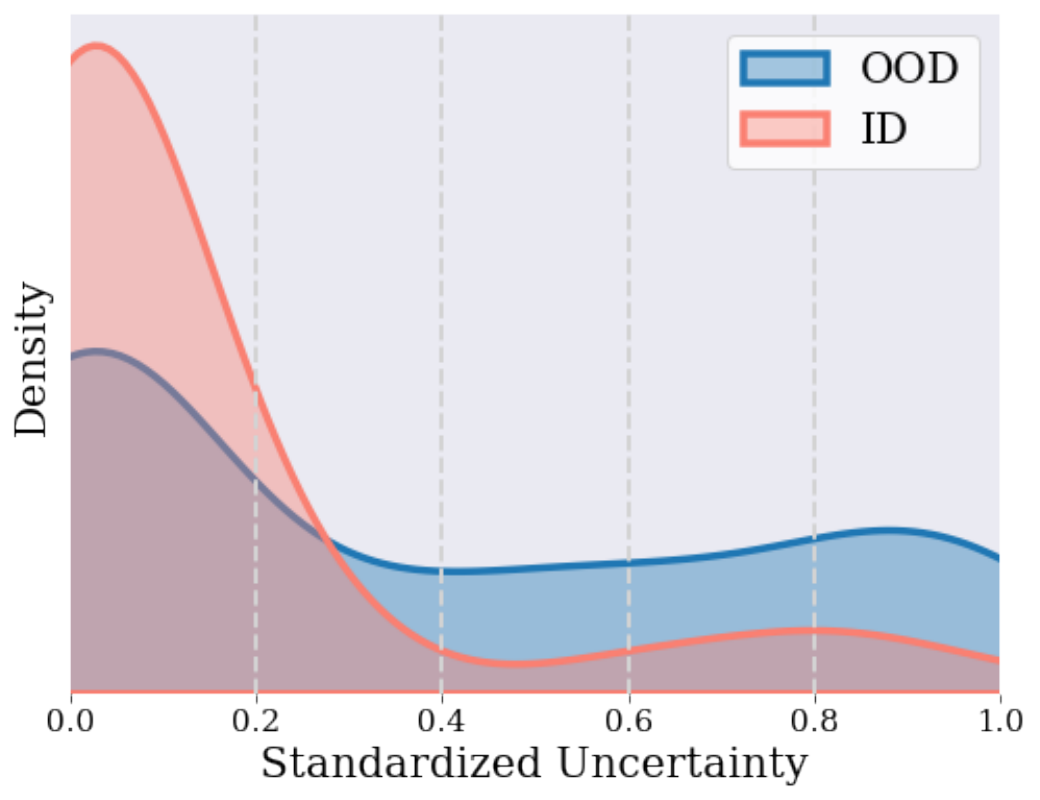}
      &
      \includegraphics[width=0.32\textwidth]{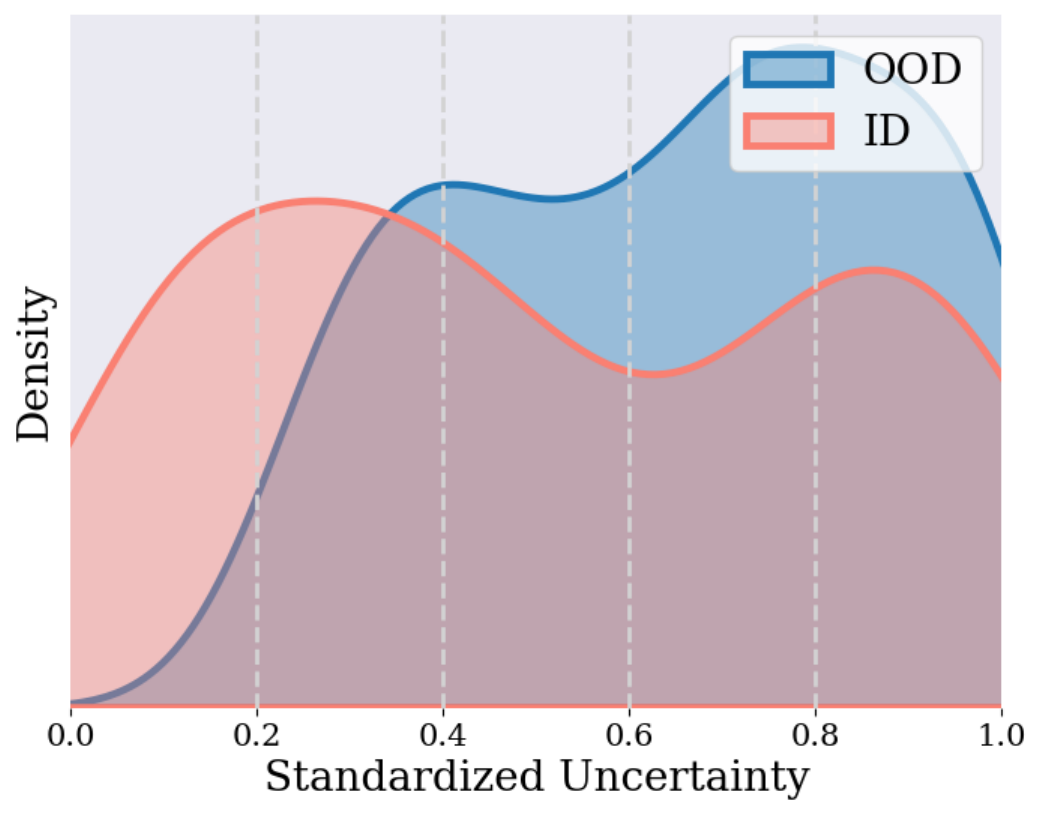}
      &
      \includegraphics[width=0.32\textwidth]{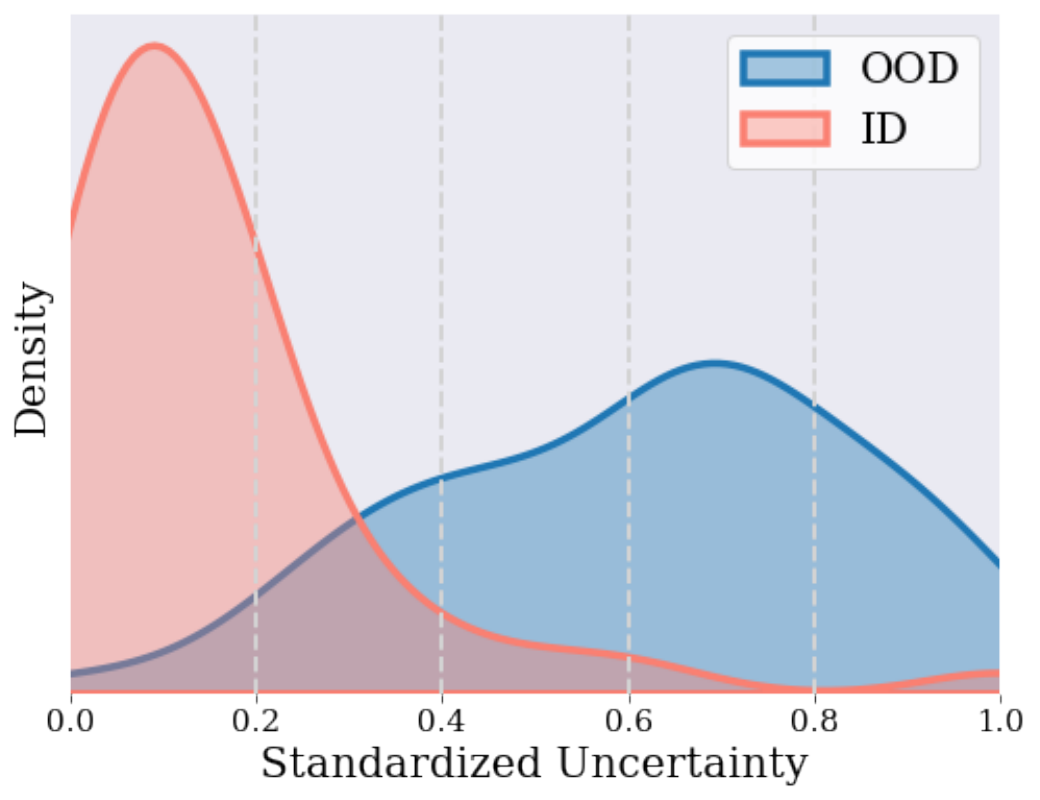}
      \\
      (d) VI & (e) PriorNet & (f) MERIT (ours)
    \end{tabular}
    \caption{Distribution of estimated uncertainty on ID and OOD Datasets by methods, including Softmax (a), Ensemble (b), Dropout (c), VI (d), PriorNet(e) and MERIT (f). MERIT demonstrated a clear distinction between the ID and OOD samples, with lower uncertainty for ID and higher uncertainty for OOD samples, which was consistent with the inherent properties of the data.}
    \label{fig:OOD_detection}
\end{figure*}

\subsection{Reliability}\label{sec:reliability}
To demonstrate reliability, we compared the proposed MERIT with other uncertainty-aware methods. 
Specifically, softmax entropy \citep{A_2021_Softmax} was selected as the baseline, and other methods estimated uncertainty using MC dropout (Dropout) \citep{C_2016ICML_dropout}, VI \citep{C_2019CVPR_Uncertainty_VI}, ensemble methods \citep{C_2021CVPR_masksembles} and PriorNet \citep{C_2018NIPS_prior_net} respectively. 
All methods utilized multi-view modeling as well for comparison, where the concatenation of local views at the input level was employed.

Specifically, we evaluated the calibration performance on a dataset without any shift in Sec. \ref{sec:cali}. 
Then we validated the capabilities of uncertainty estimation in a feature shift scenario in Sec. \ref{sec:feature_shift}. 
Moreover, we evaluated the efficacy of our class distribution-aware base rate in a class distribution shift scenario in Sec. \ref{sec:label_shift}.

\subsubsection{No distribution shift scenario}\label{sec:cali}
To quantify the discrepancy between model confidence and actual accuracy, we assessed the Expected Calibration Error (ECE) following \citet{C_2017ICML_ECE}. ECE is calculated as Eq. (\ref{eq:ECE}),

\begin{equation}\label{eq:ECE}
\mathrm{ECE}=\sum_{m=1}^M\frac{|D_m|}N\left|ACC(D_m)-{Conf}(D_m)\right|,    
\end{equation}
where $N$ samples are binned into $M$ groups and the accuracy and conﬁdence for each group are computed. $D_m$ denotes the samples whose conﬁdence predictions are in the range of $(\frac{m-1}{M} ,\frac{m}{M}]$, where the expected accuracy of the $D_m$ is $ACC(D_m) = \frac{1}{|D_m|} \sum_{i\in D_m} 1(\hat{y_i} = y_i )$, and the average conﬁdence on bin $D_m$ is calculated as $Conf(D_m) = \frac{1}{|D_m|} \sum_{i\in D_m} p(\hat{y_i} = y_i\mid x_i, \theta )$. ECE is calculated by summing up the weighted average of the differences between accuracy and the average conﬁdence over the bins. 

Table \ref{tab:uncertainty} shows that MERIT achieved a remarkable ECE, indicating a strong correspondence between model confidence and overall results. 
However, while we did not secure the top position, this could be attributed to our approach primarily estimating distributional uncertainty. In scenarios without distributional shifts, MERIT may not fully leverage its strengths.
In addition, MERIT achieved better results in terms of ACC and AUC for both tasks than the other uncertainty-aware methods. 
This suggests that the uncertainty estimated in our framework provides a clearer indication of the reliability of each view, leading to more accurate final predictions through our proposed rule-based combination scheme.

\begin{figure*}[t]
    \centering
    \begin{tabular}{cccc}
        \includegraphics[width=0.48\columnwidth]{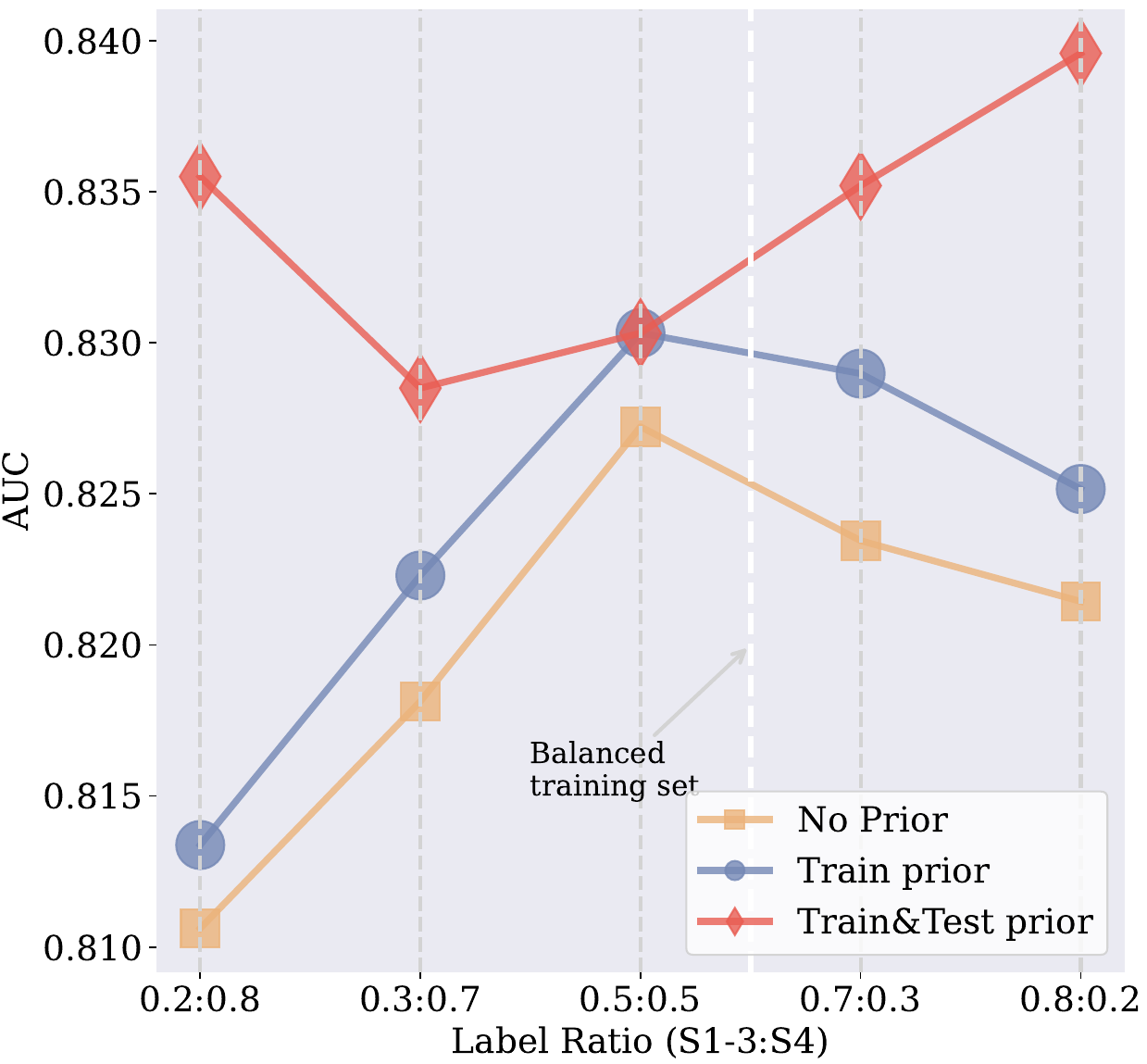} 
        & 
         \includegraphics[width=0.48\columnwidth]{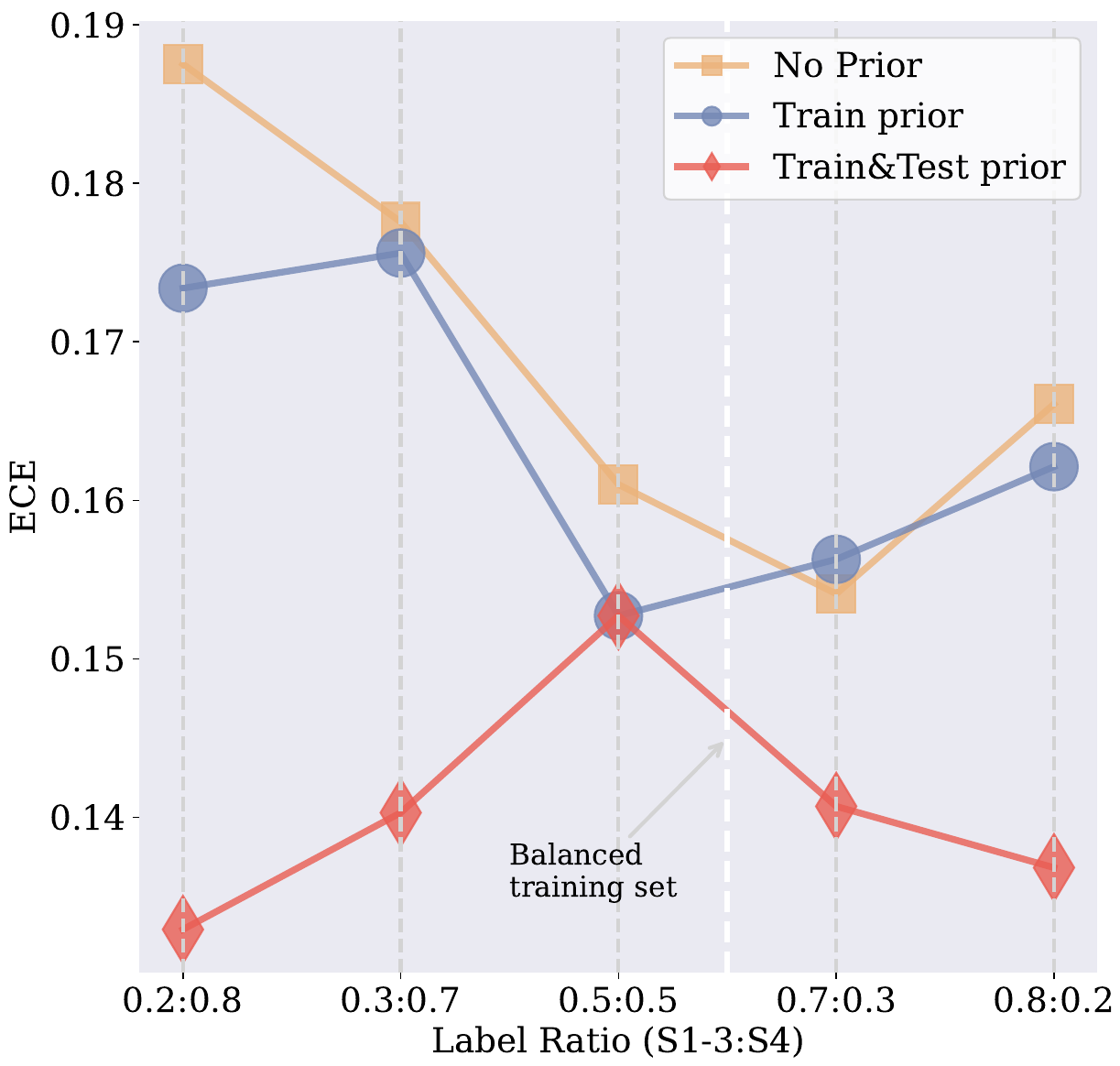}
        &
        \includegraphics[width=0.48\columnwidth]{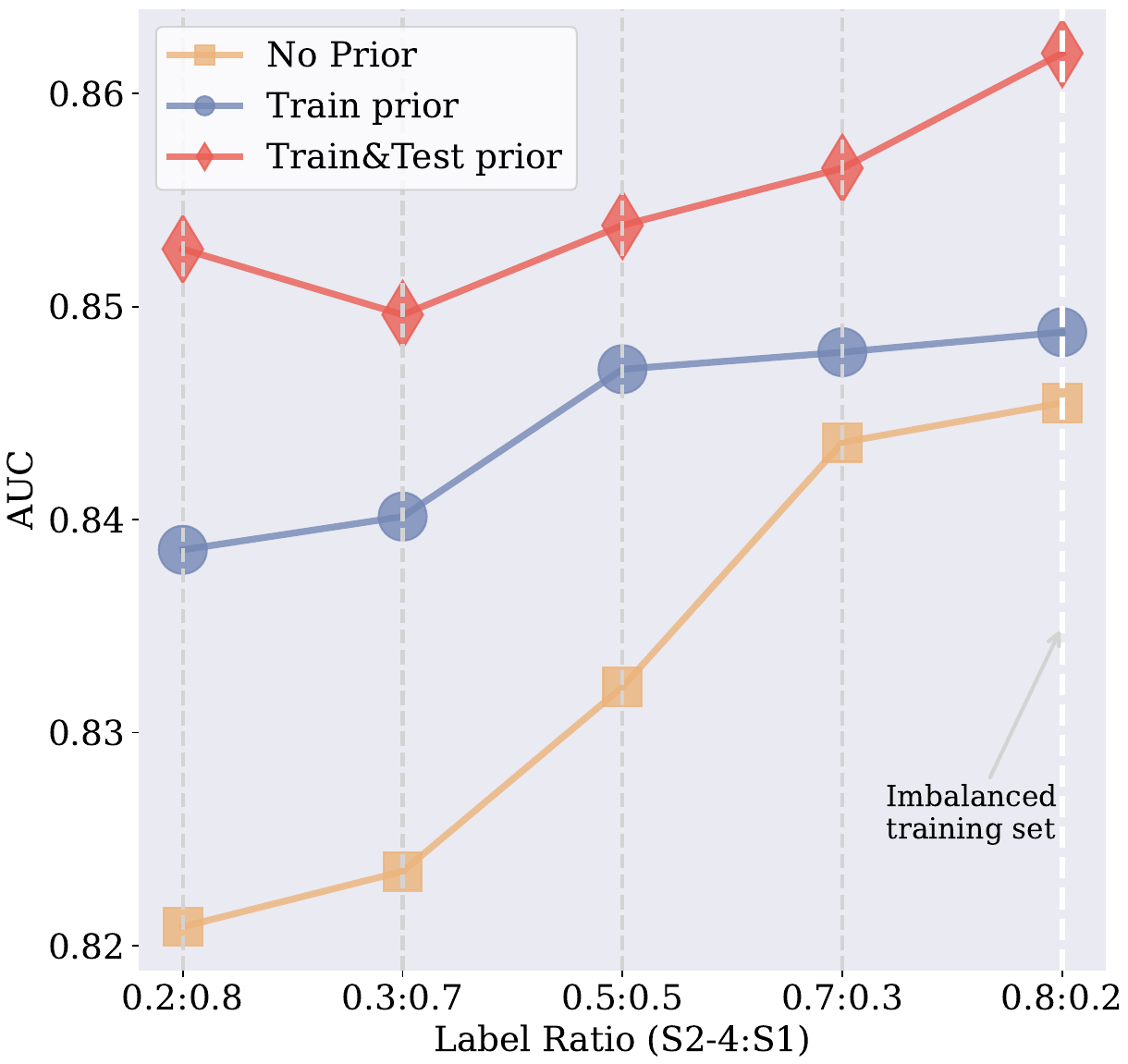}
        &
        \includegraphics[width=0.48\columnwidth]{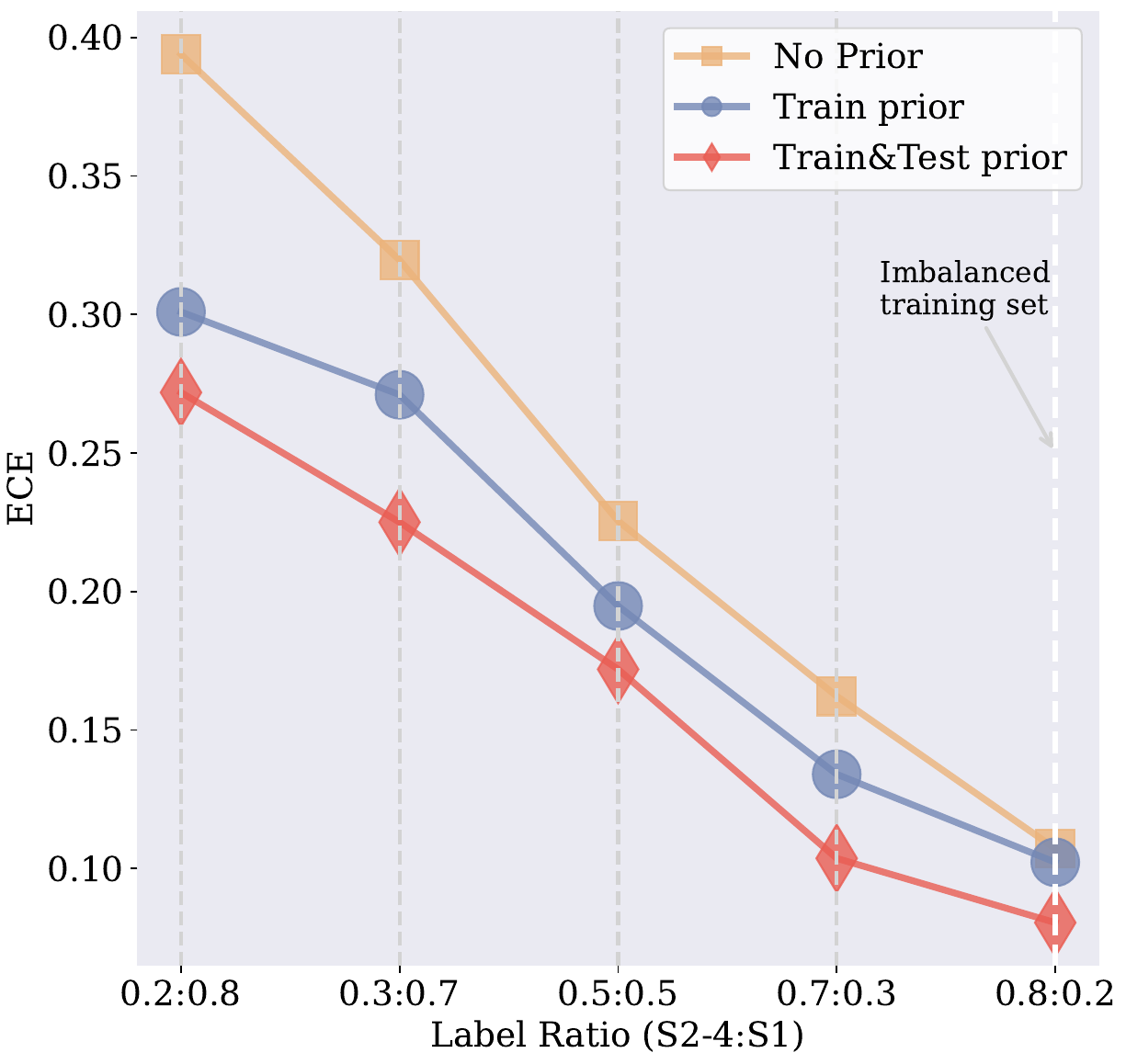}
                \\
        \multicolumn{2}{c}{(a) Balanced training set}
         &        
        \multicolumn{2}{c}{(b) Imbalanced training set}
    \end{tabular}
    \caption{Comparative analysis of AUC and ECE performance across different prior information strategies under varying class ratios. The x-axis shows the proportion of classes in the test set, and the y-axis shows the AUC and ECE metrics, illustrating the impact of prior information on the performance and reliability of the model under different class distributions. The lines represent the incorporation of prior class ratio information: yellow for \textit{No Prior}, blue for \textit{Train Prior} (introduced during training), and red for \textit{Train}\&\textit{Test Prior} (introduced during both training and testing). In (a) \textit{Cirrhosis (S4 vs S1-3)}, which has balanced training data, the introduction of test prior information notably enhances AUC performance and reduces the ECE. In (b) \textit{Substantial Fibrosis (S1 vs S2-4)}, characterized by imbalanced training data, the incorporation of train and test prior information both yield improvements over the no prior baseline.}
    \label{fig:Label-imbalance}
\end{figure*}

\begin{table*}[t]
    \centering
    \caption{Comparison with other fusion strategies. Our feature-specific method achieved the best performance.}
    \resizebox{0.9\textwidth}{!}{
    \begin{tabular}{ccccccc}
    \hline
    \multirow{2}{*}{Method} &  \multicolumn{3}{c}{Cirrhosis(S4 vs S1-3)} & \multicolumn{3}{c}{Substantial Fibrosis(S1 vs S2-4)}\\
    \cline{2-7}
    \multirow{2}{*}{}          &ACC $\uparrow$  & AUC $\uparrow$& ECE $\downarrow$  &ACC $\uparrow$&AUC $\uparrow$&ECE $\downarrow$\\
        \hline
    CBF solely   & $80.1\pm2.47$  & $85.0\pm3.11$ & $0.101\pm0.027$   & $81.9\pm2.92$ & $85.7\pm2.17$ & \bm{$0.121\pm0.023$}   \\
    BCF solely   & $79.0\pm1.64$  & $84.3\pm1.38$ & $0.188\pm0.027$   & $82.7\pm2.57$ & $85.8\pm1.76$ & $0.133\pm0.022$         \\
        \hdashline[8pt/5pt]
    MERIT (Ours) & \bm{$80.4\pm1.53$} & \bm{$85.5\pm1.54$} & \bm{$0.100\pm0.012$} & \bm{$84.1\pm2.84$} & \bm{$87.4\pm3.27$} & {$0.154\pm0.029$}  \\     
     \hline
    \end{tabular}
    }
    \label{tab:combine}
\end{table*}

\subsubsection{Feature shift scenario}\label{sec:feature_shift}
To validate the reliability in the feature shift scenario, we conducted experiments on an OOD dataset, including OOD detection and calibration. 
As shown in Fig. \ref{fig:OOD_data}, the experiment utilized OOD samples characterized by effective area, since samples with small effective area are usually considered as abnormalities in fibrosis staging \citep{J_2019R_liver_effetive_area}. 
For this purpose, liver scans with an effective area exceeding $90\%$ were categorized as the in-distribution (ID) data for training, comprising $577$ cases. Conversely, liver slices with less than $90\%$ of the area were considered as the OOD data and used for testing, consisting of $126$ cases. The ID dataset was then split into training and validation with a ratio of $8:2$. MERIT was trained on the training set. 

In the experiment of OOD detection, we utilized scaled uncertainty to recognize OOD samples. Specifically, for the quantification of uncertainty, we used Eq. (\ref{eq:opinion}) in our method, while for other methods, we followed \citet{C_2018NIPS_prior_net}, adopting the entropy with multiple sampling results to measure the uncertainty, \textit{i.e.}, $u={\mathrm{Ent}(\mathrm{output})}=-\sum_{k=1}^{K}\bar{p}_k\log \bar{p}_k$, where $\bar{p}_k$ represents the average of the $N$ probability outcomes for class $k$.  
In practical implementation, we set $N=10$. For PriorNet, $\bar{p}$ was estimated through per-data prior Dirchlet distribution, \textit{i.e.}, $\bar{p}_k=\mathbb{E}_{Dir}[p_k]=\alpha_k / \sum_k \alpha_k$. 
Then we scaled the uncertainty output into $[0,1]$ with min-max normalization across the validation and test dataset, and we established the $50\%$ percentile as the threshold, above which test samples were classified as OOD samples. 

As demonstrated in Table \ref{tab:OOD_detection}, compared to other uncertainty-aware methods, MERIT exhibited a definitive lead in the task of OOD detection, with nearly double the accuracy improvement compared to the baseline. 
This significant enhancement can be attributed to our model's reliable capability in representing distributional uncertainty to identify OOD samples effectively. 
Moreover, as illustrated in Fig. \ref{fig:OOD_detection}, MERIT differentiated between ID and OOD samples distinctly, compared to other methods. Notably, in comparison to ID samples, OOD samples exhibited a higher uncertainty, which shows MERIT can accurately estimate the distributional uncertainty.

In addition, we evaluated the performance of staging Cirrhosis (S4 vs 1-3) on the ID validation dataset and OOD dataset, to explore the capability of calibration. 
As shown in Table \ref{tab:OOD_detection}, MERIT yielded an obvious ECE increase on the OOD dataset, compared to the ID dataset. This suggests that lower uncertainty led to more accurate staging results, yielding well-calibrated performance. Meanwhile, compared to other uncertainty-aware methods, we achieved lower calibration error. 
Moreover, we observed a significant superiority in ACC and AUC for the OOD dataset, achieving $70.1\%$ and $74.1\%$, respectively. These advancements also demonstrated MERIT's exceptional generalization capability.

\begin{figure*}[t]
    \centering
        \begin{tabular}{ccc}
        \fbox{\includegraphics[width=0.3\textwidth]{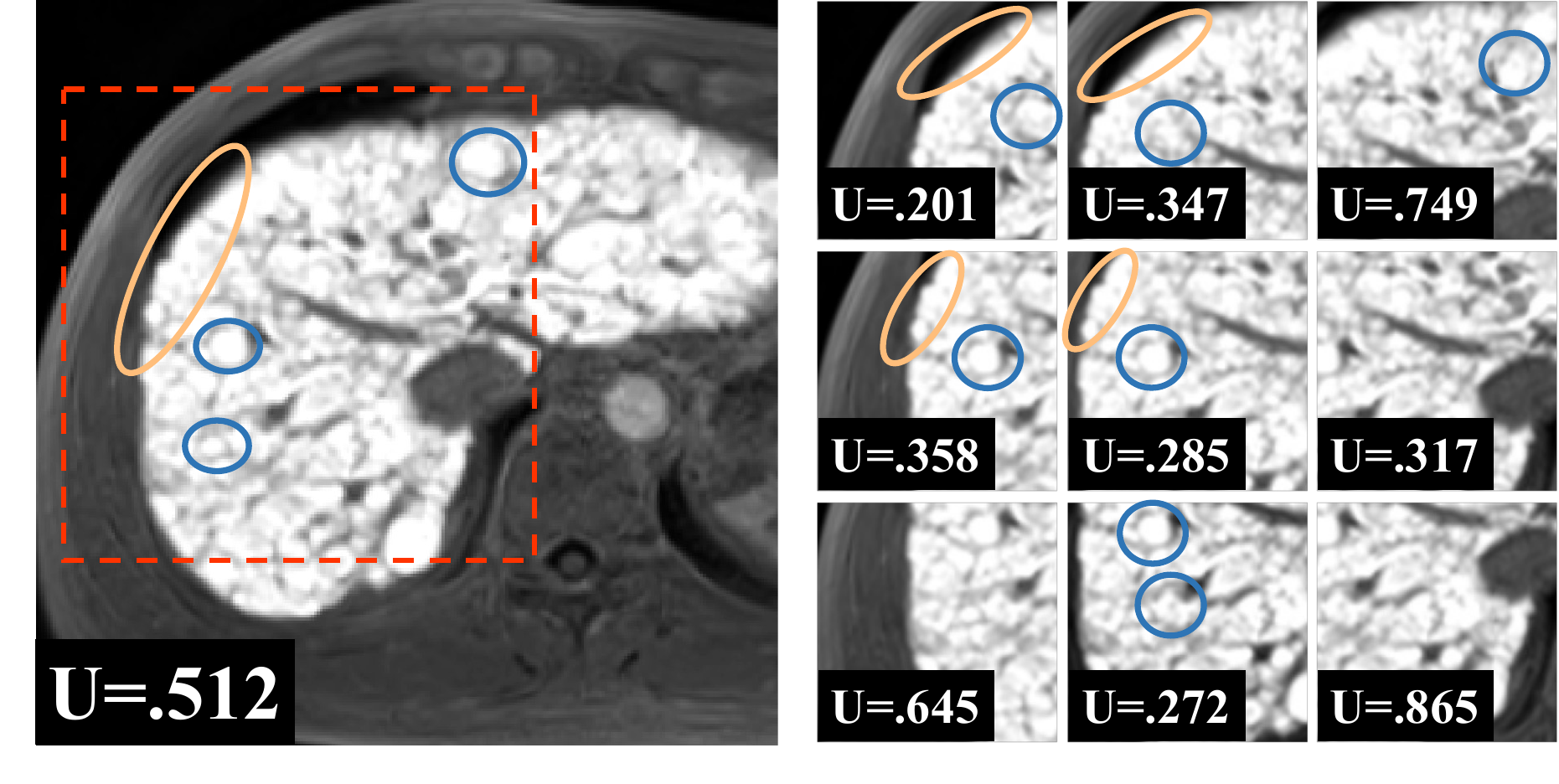}} &
        \fbox{\includegraphics[width=0.3\textwidth]{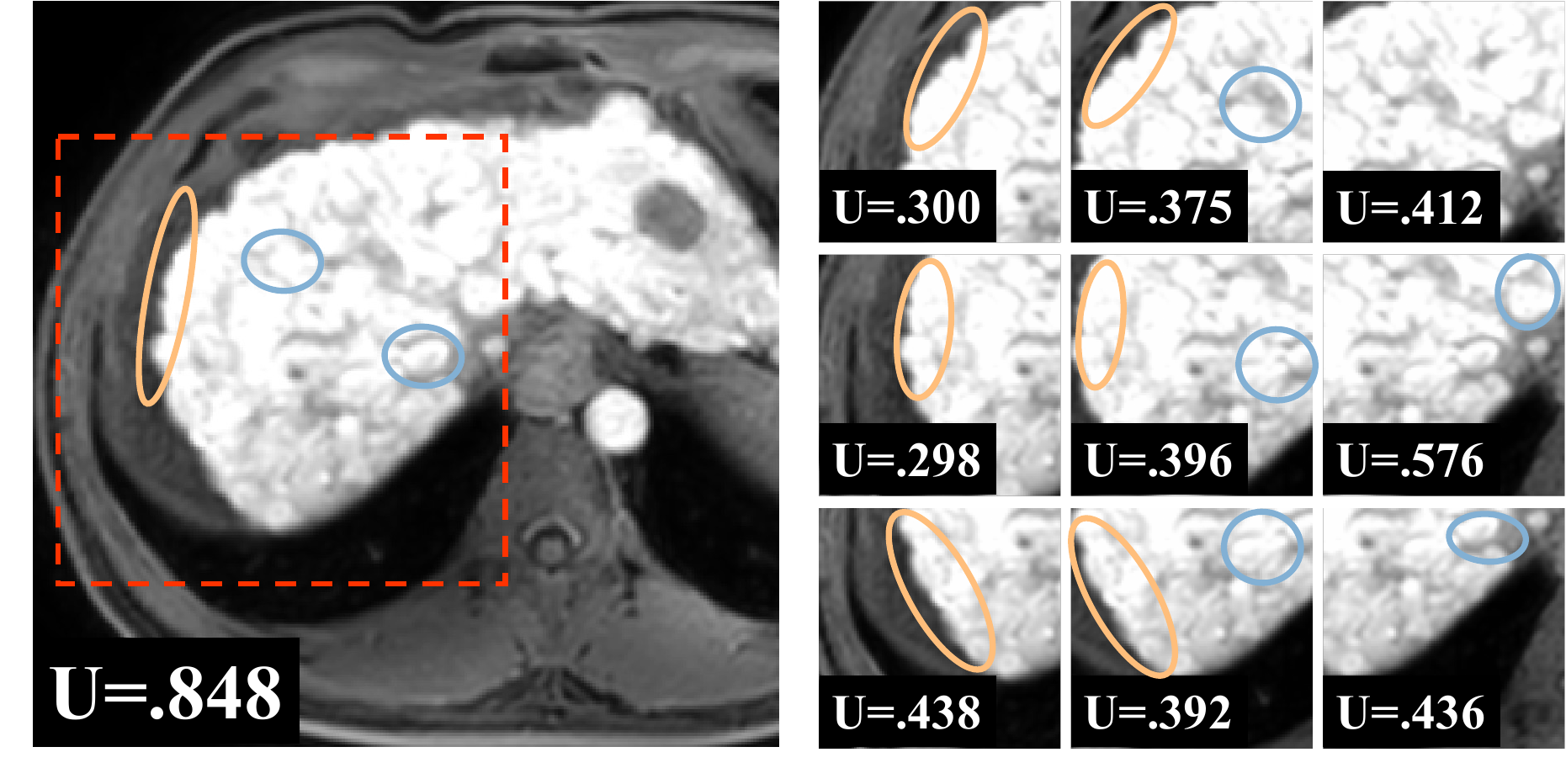}} & 
        \fbox{\includegraphics[width=0.3\textwidth]{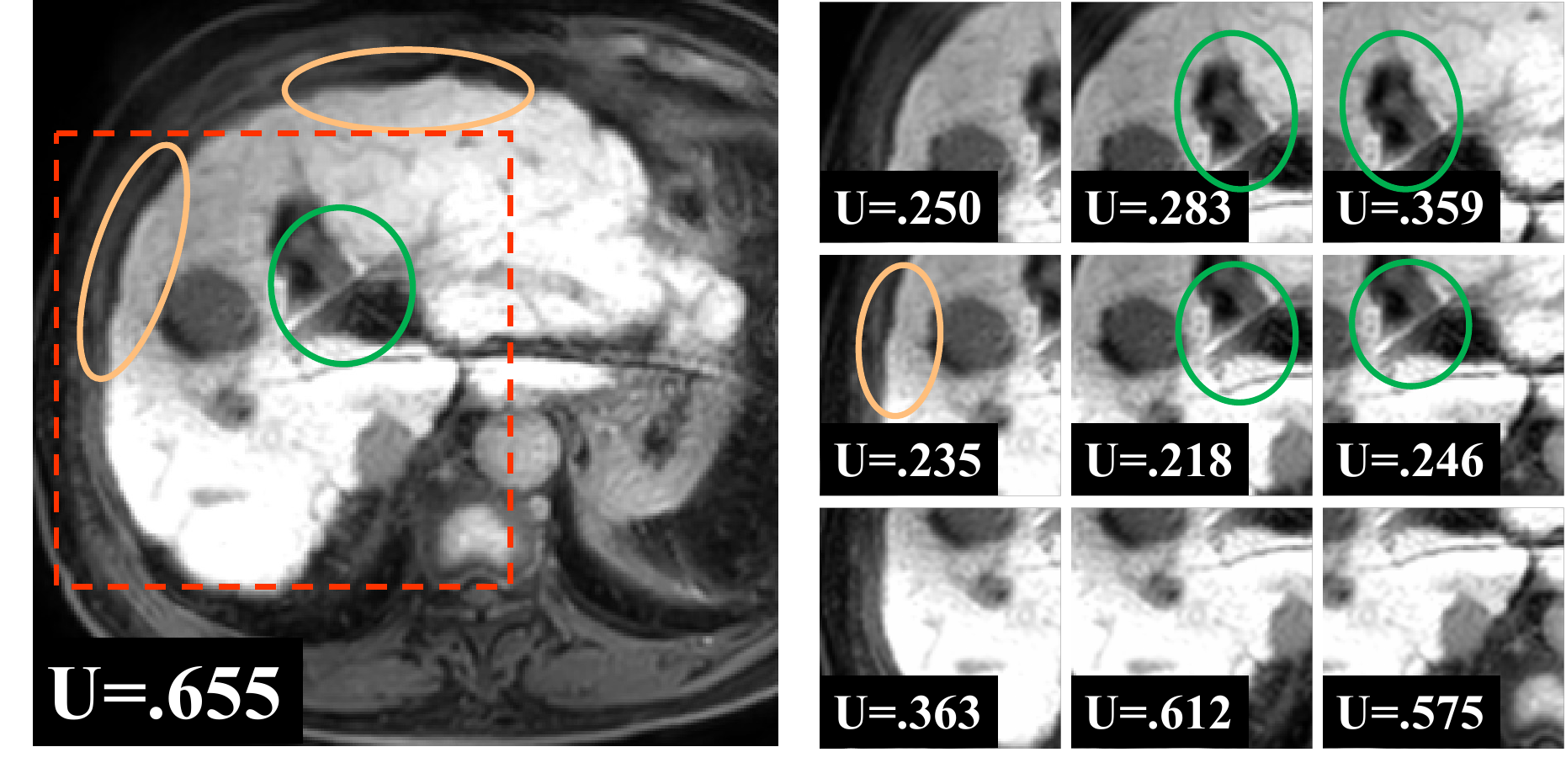}} 
        \\
        (a) & (b) & (c)
        \\
        \fbox{\includegraphics[width=0.3\textwidth]{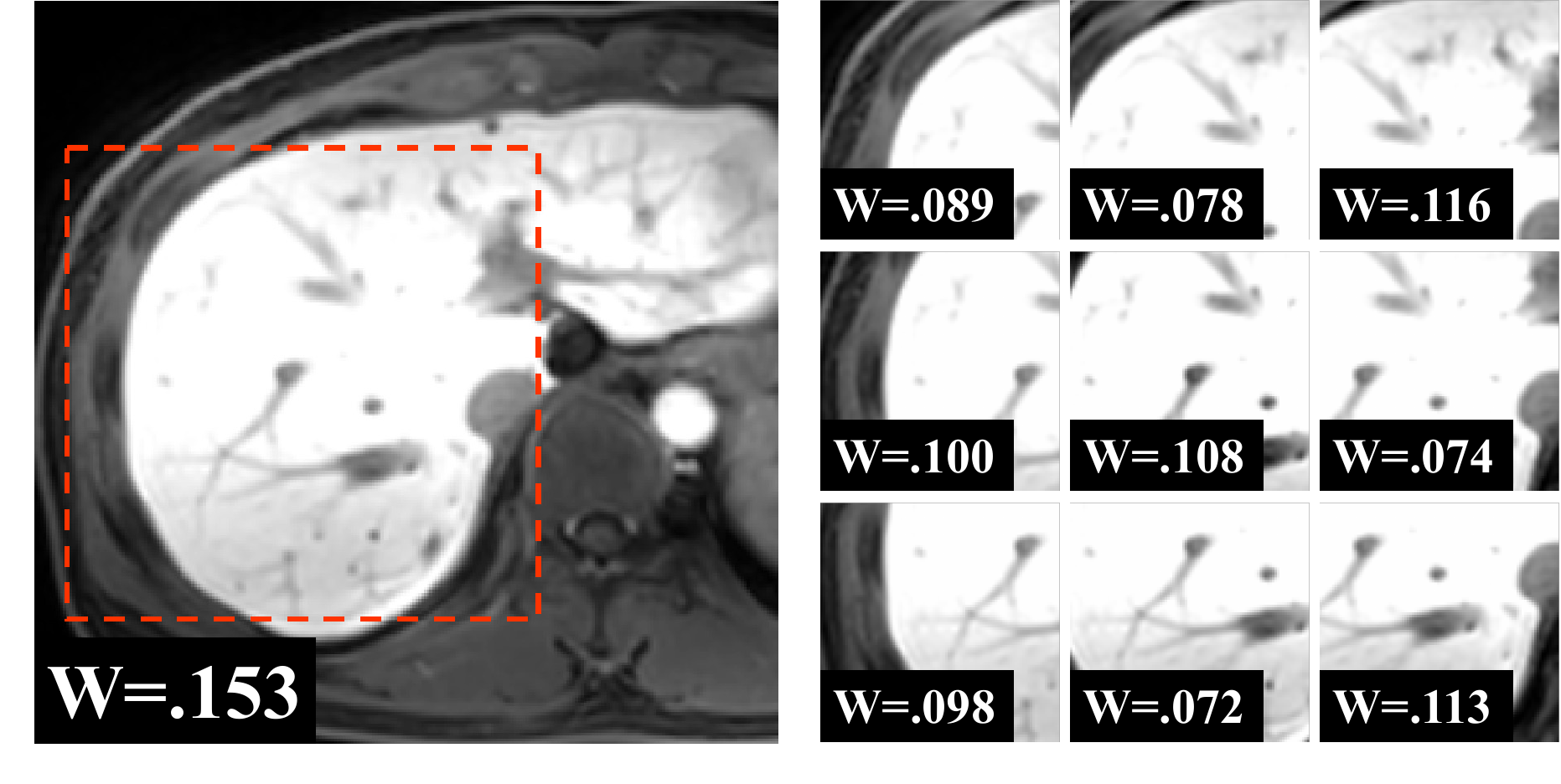}} &
        \fbox{\includegraphics[width=0.3\textwidth]{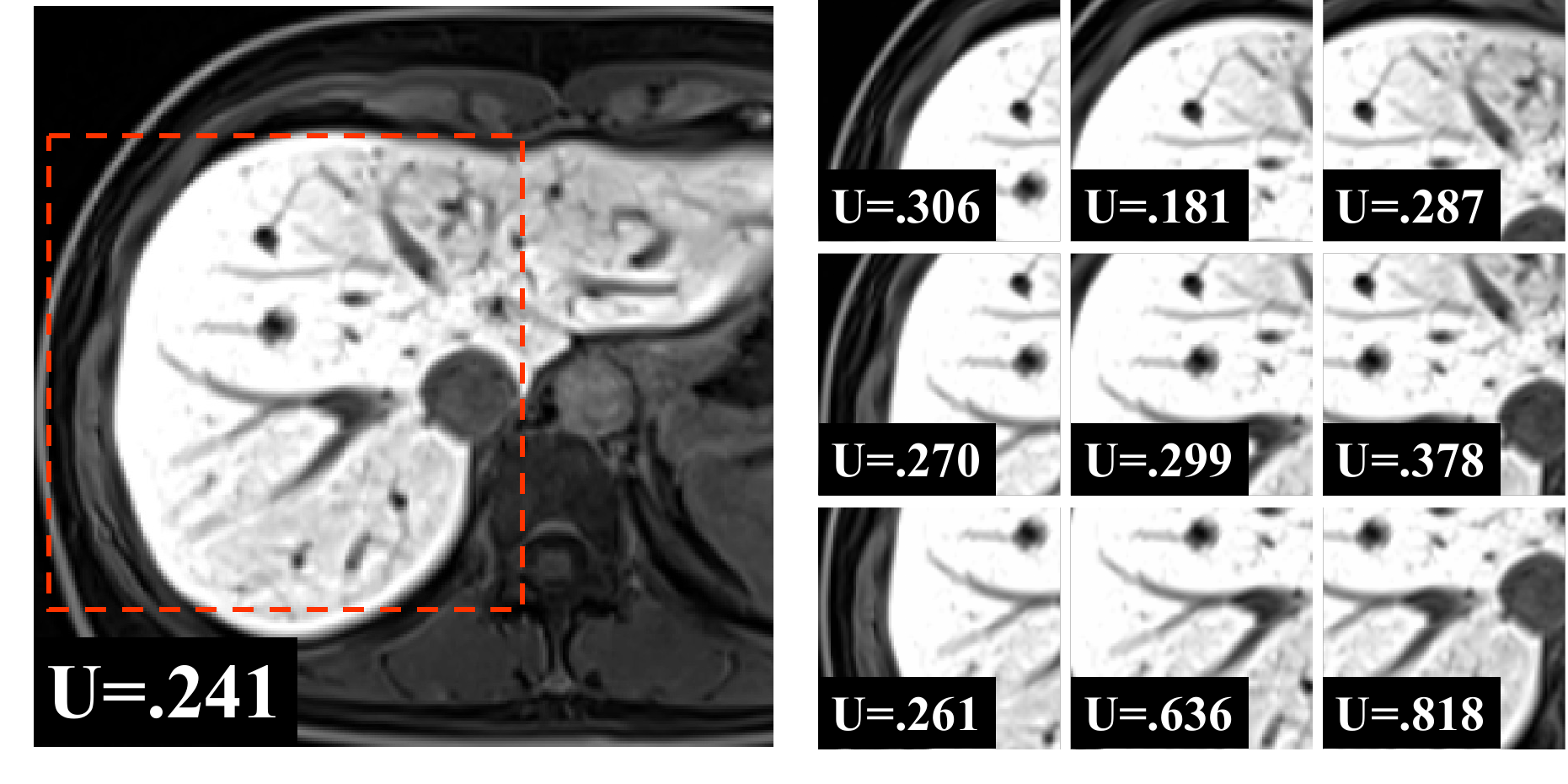}} &
        \fbox{\includegraphics[width=0.3\textwidth]{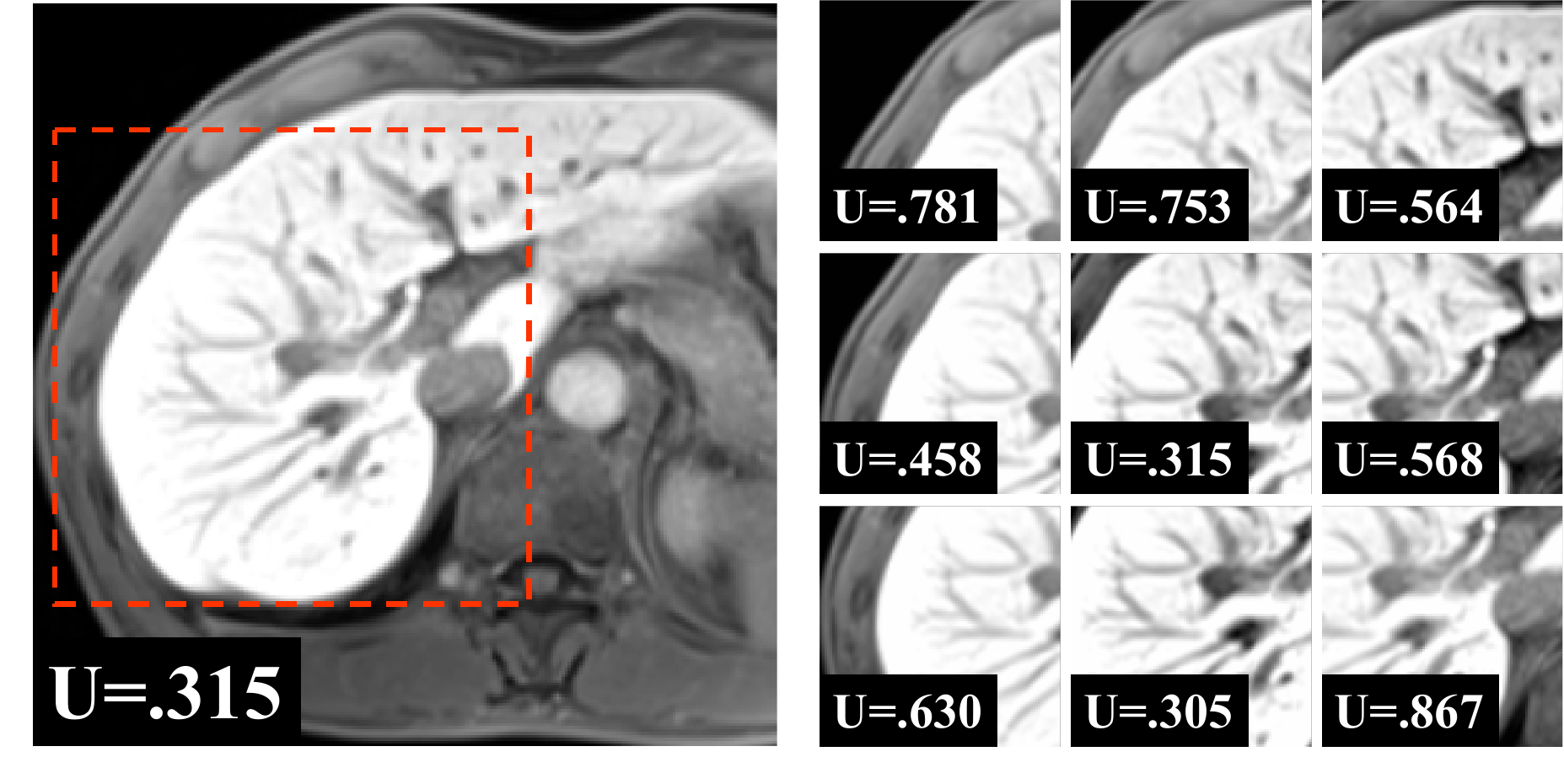}}  
        \\
     (d) & (e) & (f)
    \end{tabular}
    \caption{Typical samples of stage 4 (a,b,c) and stage 1 (d,e,f). Visible signs of liver fibrosis are highlighted by circles. Yellow circles indicate the nodular surface contour, green circles denote the widening of liver fissures, and blue circles denote numerous sclerotic regenerative nodules. Uncertainties (U) of local and global views estimated by our model were demonstrated. Notably, local views of lower uncertainty contain more signs of fibrosis. Please refer to supplementary materials for more high-resolute images.}
    \label{fig:signs}
\end{figure*}

\subsubsection{Class distribution shift}\label{sec:label_shift}
To validate the effectiveness of the class distribution-aware base rate, we performed ablation studies in the class distribution shift scenario. 
The class distribution shift was simulated by under-sampling the test set to achieve specified class proportions, ranging from $2:8$ to $8:2$. 
In each proportion, we compared three methods: 
(1) No Prior: No prior category information was incorporated during both training and testing; 
(2) Train Prior: Prior information was included during training solely; 
(3) Train\&Test prior: Prior information was incorporated during both the training and testing process. 
Since under-sampling was performed on the test set, cross-validation was not used in this experiment. Instead, we randomly split the dataset into training, validation, and test in a $6:2:2$ ratio and repeated this process three times in each setting across the three methods. AUC and ECE were employed for evaluation.

The results depicted in Fig. \ref{fig:Label-imbalance} offer insightful observations regarding the impact of class distribution prior information on model performance across the three methods. 
In Fig. \ref{fig:Label-imbalance}(a), which shows the experiment results on the balanced training dataset, the introduction of the class prior information showed varied impacts on performance. 
Notably, the yellow and blue lines, representing strategies without prior and with training prior, exhibit tiny differences, which could be attributed to the inherent balance in class distribution of training set. 
However, the red line, representing the strategy where prior information was introduced during both training and testing, shows a significant boost in performance. This enhancement is particularly evident in highly imbalanced scenarios, suggesting that the introduction of the informative base rate allows for a post-hoc correction during testing that can significantly improve the model's predictive accuracy. 

In Fig. \ref{fig:Label-imbalance}(b), when considering the imbalanced training dataset scenario, there is a performance improvement ($22.7\%$ relatively in ACC of minority class with uniform test set) when training prior information was integrated into the learning process, compared to the strategy without any prior information. 
This suggests that by incorporating class ratio information during training, the feature extracted by the network will not be biased toward the majority class, thereby improving the performance of the minority class.

\subsection{Study of Interpretability} 
In this subsection, we first validated our assumptions about the design of interpretable combination rules by comparing the performance of different configurations. 
Then we evaluated the quality of post-hoc interpretation through several case studies.

\subsubsection{Feature-specific combination rule} \label{sec:rule}
To evaluate the effectiveness of interpretable combination rules for integrating global and local views, 
we conducted comparison experiments on three distinct configurations: both global and local views using the CBF operator, both views employing the  BCF operator, and a hybrid approach where the local views were combined using the CBF operator, while the integration of local and global views was based on the BCF operator. 

Table \ref{tab:combine} presents a comparative analysis of fusion strategies for view integration. 
It could be observed that our method could yield consistently better results than the method that only uses the BCF operator.
This validates the usage of the CBF operator for the combination of local views.
When compared with the method using CBF solely, our method achieved $2.2\%$ and $1.7\%$ improvement in ACC and AUC of the substantial fibrosis identification task, respectively, which could demonstrate the effectiveness of using BCF for the combination of the global and local views.
Although CBF performed slightly better than our method in ECE of the substantial fibrosis task, the calibration performance was comparable in the cirrhosis staging task.
This could be due to the fact that using CBF solely would result in the addition of evidence and then yield low uncertainties. Therefore, its calibration performance would be better in the substantial fibrosis task, where the accuracy was high.

\subsubsection{Post-hoc interpretation} \label{sec:post}
MERIT mainly provides post-hoc interpretation at the prediction level by measuring the feature importance through uncertainties \citep{J_2021_feature_interpret}.
To demonstrate that the predicted uncertainties could discern the contribution of individual views to the final decision, we presented visual results of typical samples from stages 4 and 1, obtained by the model for staging cirrhosis.

In Fig. \ref{fig:signs}, the colored circles, annotated by experienced doctors, highlight the signs of fibrosis. 
Fig. \ref{fig:signs}(a,b,c) demonstrated samples from stage 4. It could be observed that the estimated uncertainty in each view is consistent with the signs of fibrosis, \textit{i.e.}, views with more signs achieve lower uncertainty, playing a more significant role in the final prediction. 
Moreover, the uncertainty derived from the global view is high, even if there are many signs of fibrosis, which suggests that the local views are primarily crucial in staging cirrhosis. 
Conversely, Fig. \ref{fig:signs}(d,e,f), which illustrate S1 samples without any fibrosis signs, reveal lower uncertainty in the global view, indicating that it is easier to make decisions from the global view in the absence of visible fibrosis signs. 

\begin{figure*}[t]
\centering
\includegraphics[width=\textwidth]{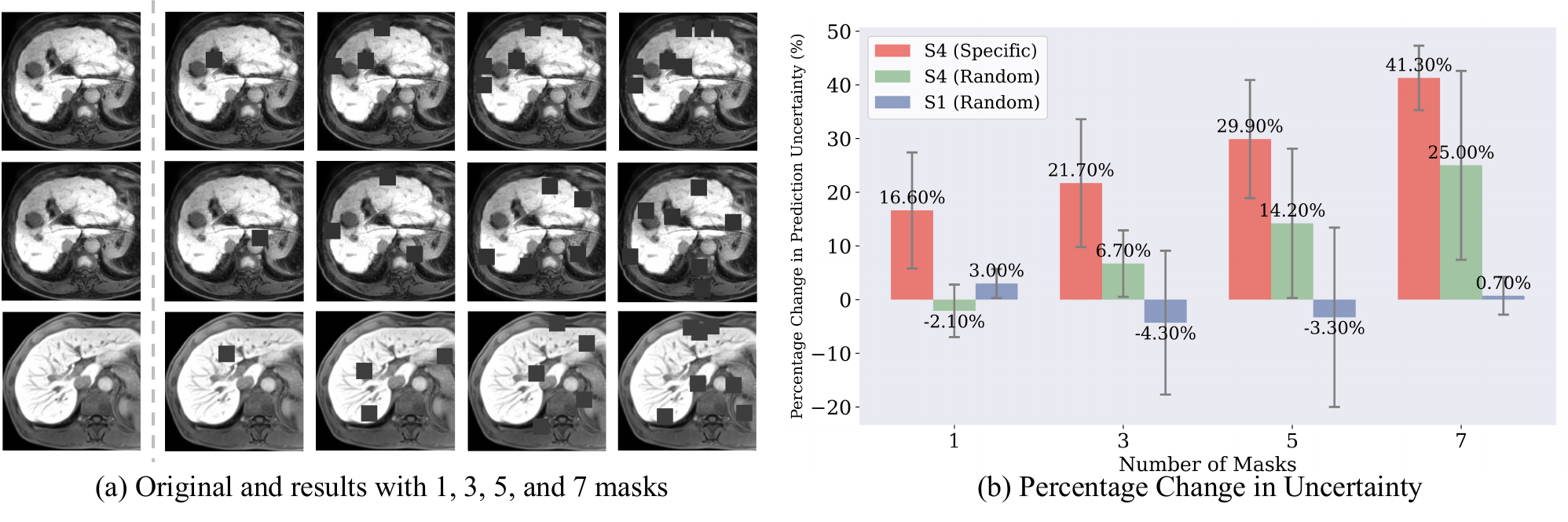}
\caption{Illustration of feature ablation experiments: {(a) The three rows represent the S4 sample with specific masks targeting fibrosis features, the S4 sample with random masks, and the S1 sample with random masks, respectively. (b) Relative percentage changes in prediction uncertainty with varying numbers and types of masks for liver fibrosis stage 4 and 1.} The results demonstrated the mean percentage change and standard deviation in prediction uncertainty when 1, 3, 5, and 7 masks were applied.}
\label{fig:mask}
\end{figure*}

To further validate that our model could explore the role of each sub-view in the final decision, we followed the idea of feature ablation \citep{C_2022NIPS_feature_ablation} and applied it to the sub-views. Specifically, we masked part of the input to study the resulting changes in uncertainty output. 
{As illustrated in Fig. \ref{fig:mask}(a), for stage 4 samples with distinct fibrosis features, we applied specific masks over these areas and used random masks as a comparison. For stage 1 samples, masking was applied randomly.}
We applied masks to $1, 3, 5$, and $7$ square areas of $30\times 30$ pixels, respectively, in both scenarios, and tested $3$ samples in each.

{As shown in Fig. \ref{fig:mask}(b), for samples in stage 4 with specific masks, the final uncertainty output increased with the increase in number of masked areas. In contrast, for cases with random masks, the increase was much slower and exhibited greater variation. This further confirms that our model effectively assessed the significance of each sub-view.}
Specifically, the sub-views with more signs of fibrosis played a more important role in the decision-making process. 
Notably, we have utilized a cutout data-augmentation technique, so that the increase of uncertainty was caused by the absence of critical fibrosis information, instead of the random mask. 
For stage 1 samples, the final uncertainty remained relatively stable, since stage 1 samples did not contain critical information. 

\subsection{Multi-view modeling} \label{sec:multiview}

\begin{table*}[t] 
    \centering
    \caption{Comparison with multi-view learning methods. Results were evaluated in accuracy (ACC) and area under the receiver operating characteristic curve (AUC) for both tasks.}
    \resizebox{0.8\textwidth}{!}{
        \begin{tabular}{ccccc}
        \hline
        \multirow{2}{*}{Method} &\multicolumn{2}{c}{Cirrhosis(S4 vs S1-3)} & \multicolumn{2}{c}{Substantial Fibrosis(S1 vs S2-4)}\\
        \cline{2-5}
        \multirow{2}{*}{}             &ACC $\uparrow$  & AUC $\uparrow$  &ACC $\uparrow$&AUC $\uparrow$ \\
        \hline
    {SingleView} \citep{J_2021ER_subview_liver}     & $69.7\pm2.55$ & $73.0\pm1.70$  &$69.7\pm3.09$ & $72.2\pm6.53$  \\   
    {DLCT}\citep{J_2018ER_liver_CT_DL}      & {{$75.0 \pm 5.29$}}
 & {$79.6\pm5.83$}  & {{$74.4\pm2.37$}} & {$77.9\pm2.21$} \\        
 {Residual} \citep{J_2020_liver_resnet} & {{$74.1 \pm 2.54$}}
 & {$77.4\pm2.63$}  & {{$70.5\pm1.49$}} & {$75.4\pm1.23$} \\    
        \hdashline[8pt/5pt]
    {Concat} \citep{C_2016CVPR_input_fusion}        & $74.4\pm1.95$ & $75.7\pm1.63$  & $74.7\pm2.41$ & $76.2\pm5.02$   \\
     {DCCAE} \citep{C_2015ICML_DCCAE}                & $75.9\pm3.03$ & $80.1\pm3.59$  & $72.2\pm2.98$ & $76.1\pm2.86$   \\
    {CMC} \citep{C_2020ECCV_Contrastive}            & $78.3\pm3.68$ & $81.5\pm3.80$  & $75.1\pm2.89$ & $77.8\pm3.04$   \\
    {PredSum} \citep{C_2014NIPS_decision_fusion_1}    & $77.0\pm3.32$ & $80.9\pm0.03$  & $74.3\pm2.99$ & $77.3\pm0.04$   \\
    {Attention} \citep{C_2018ICML_attention_fusion} & $73.1\pm2.25$ & $75.3\pm0.02$  & $71.3\pm3.11$ & $73.8\pm0.04$   \\   
        \hdashline[8pt/5pt]
    {Local views solely}  & $75.8\pm1.70$ & $82.2\pm0.99$  & $83.7\pm1.75$  & $83.4\pm5.23$\\   
    {Global views solely} & $75.4\pm3.64$ & $77.1\pm4.33$  & $74.4\pm2.79$ & $75.6\pm5.52$   \\   
    {Both with CNN} & $80.0\pm2.02$ & $85.3\pm2.11$ & $82.7\pm0.47$ & $85.3\pm2.55$   \\   
    {Dilated CNN for global view} & $79.4\pm 3.31$ & $84.6\pm3.72$ & $82.4\pm1.10$ & $86.2\pm4.39$   \\   
            \hdashline[8pt/5pt]
         {MERIT(w/o)~\citep{C_2023MICCAI_ours}} & {{$79.6 \pm 3.39$}}  & {$82.3\pm1.69$}  &{{$82.9\pm1.91$}} & {$82.4\pm2.01$} \\     
MERIT (Ours)     & \bm{$80.4\pm1.53$} & \bm{$85.5\pm1.54$}& \bm{$84.1\pm2.84$} & \bm{$87.4\pm3.27$}  \\        
        \hline
        \end{tabular}
        }
        \label{tab:multiview}
\end{table*}

To assess the effectiveness of the proposed MERIT, we compared it with five multi-view learning methods, including Concat \citep{C_2016CVPR_input_fusion}, DCCAE \citep{C_2015ICML_DCCAE}, CMC \citep{C_2020ECCV_Contrastive}, PredSum \citep{C_2014NIPS_decision_fusion_1}, and Attention \citep{C_2018ICML_attention_fusion}. 
Concat is a commonly used method that concatenates multi-view images at the input level. DCCAE and CMC are feature-level strategies. PredSum and Attention are based on decision-level fusion.
{In addition, we included comparisons with three single-view methods designed for liver fibrosis staging: SingleView \citep{J_2021ER_subview_liver}, which uses a single patch as input, and DLCT~\citep{J_2018ER_liver_CT_DL} and Residual~\citep{J_2020_liver_resnet}, both of which only rely on the whole image as single input.}
In addition, we also performed ablation studies to validate our incorporation of the global view, where a data-efficient transformer was employed to extract features. Specifically, we compared methods that use local views solely, global views solely, and both views with CNN. Moreover, we employed an alternative method to capture long-range dependency of the global view feature, using dilated CNN \citep{C_2016ICLR_dilated}, where we set the dilation rate as $1,1,2,4$ in four blocks, respectively.
We have also included our previous MICCAI version, denoted as MERIT(w/o), to highlight the effectiveness of the improvements introduced in this work, specifically the distribution-aware base rate to address the label-imbalance problem and feature-specific fusion operators to faithfully model the relationship between local and global view features.

Compared with multi-view learning methods, the results summarized in Table \ref{tab:multiview} demonstrate that MERIT significantly improves performance over the SingleView method by $12.5\%$ and $15.2\%$ in AUC on two different tasks.
This indicates that MERIT can extract and utilize features more effectively than those that rely on a single view. 
Moreover, MERIT outperforms two methods (DLCT and Residual) that use only the whole image as input, showcasing its capability to leverage fine-grained features from local views and enhance final decision fusion.
Additionally, our model outperforms other multi-view learning approaches in staging liver fibrosis, likely due to its enhanced capability to capture a broader range of features, encompassing both global and local perspectives. Furthermore, our feature-specific fusion strategy proves to be more effective than conventional methods. 
Specifically, we analyze the weights assigned to each view during the fusion process in attention-based fusion method~\citep{C_2018ICML_attention_fusion}.
The results revealed that, in comparison to our method, the attention-based method did not effectively differentiate between views, failing to distinguish the contributions of each view to the final decision. This indicates a limitation in identifying and prioritizing useful information over non-useful details. The detailed visualization is provided in Fig. \ref{fig:weight_compare} in supplementary material.
Compared to MERIT(w/o), our method achieves superior performance in both tasks, demonstrating the effectiveness of the introduced improvements. Specifically, in the more imbalanced Substantial Fibrosis task, our method improves AUC by $6.07\%$.

Notably, in the ablation study, using the global view solely achieved the worst performance, indicating that it could be challenging to extract useful features without any complementary information from local views. In addition, compared with the method that only used local views, MERIT also gained improvement. 
Meanwhile, MERIT also performed better than the method that applied a convolution neural network for the global view, even when using dilated convolution which has a larger receptive field. This demonstrates that the proposed data-efficient transformer was more suitable for the modeling of global representation than CNN. 

\section{Discussion}
\subsection{Clinical Significance}
{The proposed model could improve the diagnostic confidence and treatment planning in practice due to the interpretable design of the combination rules and the ability to quantify distributional uncertainties. 
First, the diagnostic confidence could be increased by integrating opinions from physicians and models \citep{J_2024_confidence}. For example, if the model gives a prediction with high uncertainty, the doctors could easily trace back to the prediction of each local view and check whether the model yields wrong predictions. The model could generate a new diagnosis result after correcting the misclassified local view.
Besides, the uncertainty predicted by our model could avoid overtreatment in treatment planning.
For instance, if the model suggests severe fibrosis with a high uncertainty score, the clinician may choose to conduct additional tests, gather more information, or consult with other experts before proceeding with the treatment, which could prevent excessive intervention that has side effects \citep{J_2024_overtreat}.
}

\subsection{Limitation and future work}
{In this study, we analyzed several misclassification cases with the help of expert clinicians, which led to some valuable insights.
As shown in Table \ref{tab:incorrect_examples} in supplementary material, we observed the following key findings:
(1) Limited Use of Single-Slice Information: Although MERIT imitates clinician diagnosis using only single-slice data to enhance the interpretability, a 3D image can provide critical information, such as liver volume reduction, that is essential for accurate staging. Recent studies \citep{J_2024NSR_3d_liver_ct} have demonstrated that fully-automated liver fibrosis staging can be performed using 3D CT data. Therefore, extending the MERIT framework to 3D liver MRI could enhance its performance, as the method could automatically distinguish between useful and non-useful slices.
(2) Potential of Multi-Modal Data: Currently, our model uses image data alone. However, integrating other modalities, such as clinical records and assessment scales, could provide additional information that influences the diagnosis. For example, patient history, such as previous surgeries, could significantly help the model in more accurately predicting liver fibrosis stages.
(3) Impact of Artifacts on Predictions: We also observed that image artifacts can degrade prediction accuracy. 
These artifacts, such as respiratory motion, can obscure important features. Future work could explore modeling these artifacts to reconstruct high-quality images and improve the model's robustness. These observations suggest directions for future improvements, including expanding the data input sources, integrating advanced imaging modalities, and refining artifact handling to enhance the model’s diagnostic accuracy.}

\section{Conclusion}\label{sec:conclusion}
In this work, we propose a reliable and interpretable multi-view learning framework for liver fibrosis staging (MERIT) based on evidential theory. Specifically, we model the liver fibrosis staging as a multi-view learning task, with cropped local image patches and a global view. Then we employ the subjective logic theory to estimate the distributional uncertainty of prediction in each view. Furthermore, we utilize a feature-specific combination rule to fuse the predictions explicitly. Additionally, we incorporate a class distribution-aware base rate to tackle the distribution shift problem. 
Quantitative and qualitative experimental results on reliability and multi-view modeling, as well as the ad-hoc and post-hoc interpretability, have shown the effectiveness of the proposed MERIT.

\section*{Acknowledgement}
The authors would like to thank Xin Gao for the valuable comments and proofreading of the manuscript.

\bibliographystyle{model2-names.bst}\biboptions{authoryear}
\bibliography{strings,refs}

\clearpage
\section*{Supplementary Material}
In the supplementary material part, we provide additional details and analyses that complement and expand upon the main findings presented in this paper. 

\subsection{Mapping between Dirichlet distribution and Opinions}
From the respective of opinion, given opinion $D=\{\{b_k\}_k=1^K,u\}$ and base rate $\bm{a}$, the projected probability $\bm{p}$ of a opinion about is deﬁned by Eq. (\ref{eq:opinion_map}) following \citet{B_2016_subjective},

\begin{equation}\label{eq:opinion_map}
    \bm{p} = \bm{b} + \bm{a} u.
\end{equation}

From the perspective of Dirichlet distribution, the point estimation of MLE for multinomial distr is the expected probability derived as Eq. \ref{eq:dir},

\begin{equation}\label{eq:dir}
  \mathbb{E}_{{Dir}\left(\bm{\mu} \mid \bm{\alpha}\right)}\left[\bm{\mu}\right]
  =\frac{\bm{\alpha}}{S}=\frac{\bm{\alpha}-\bm{a}{W}+ \bm{a}{W}}{S}
  =\frac{\bm{\alpha}-\bm{a}{W}}{S}+\bm{a}\frac{{W}}{S}.
\end{equation}

The bijective mapping between Dirichlet distribution and opinions is based on the requirement for equality between the projected probability $\bm{p}$ derived from opinion $D$, and the expected probability distribution $\mathbb{E}_{Dir}$ derived from Dirichlet distribution $Dir(\bm{p}\mid \bm{\alpha})$. This requirement is expressed as $\bm{p}=\mathbb{E}_{Dir}$. By combining Eq. (\ref{eq:opinion_map}) and (\ref{eq:dir}), we obtain the following result, 

\begin{equation}
    \begin{aligned}
        \bm{b}+\bm{a}u =\frac{\bm{\alpha}-\bm{a}{W}}{S}+\bm{a}\frac{{W}}{S}.
    \end{aligned}
\end{equation}

To proceed with the derivation, we employ two intuitively reasonable assumptions in \citet{B_2016_subjective}: (1) an increase in evidence $e_k$ enhances belief in its corresponding output $b_k$, (2) the accumulation of total evidence $\sum_{k=1}^Ke_k$ reduces the uncertainty $u$. Based on these two assumptions, we can then derive the mapping, 

\begin{equation}
    \bm{b} = \frac{\bm{\alpha}-\bm{a}{W}}{S}, u = \frac{{W}}{S},
\end{equation}

\subsection{Derivation of Equations}
\noindent
\textbf{Equation (\ref{eq:add_evidence})}: According to Eq. (\ref{eq:evidence}) and Eq. (\ref{eq:opinion}), the evidence of the combined opinion $\bm{D}=\bm{D}^m \oplus \bm{D}^n$ could be written as,
\begin{flalign}\label{eq:evib}
    \bm{e} = \frac{K}{u}\bm{b}.
\end{flalign}
For each class $k$, combining Eq. (\ref{eq:evib}) and Eq. (\ref{eq:CBF}), we have,
\begin{flalign}
    \bm{e}_k &= \frac{K(b_k^m u^n + b_k^n u^m)}{u^m u^n} \nonumber\\
    &= \frac{Kb_k^m}{u^m} + \frac{Kb_k^n}{u^n}=\bm{e}_k^m + \bm{e}_k^m
\end{flalign}

\vspace{0.5cm}
\subsection{Proof of propositions}
\setcounter{proposition}{0}
\begin{proposition}
    If both opinions are equally confident, \textit{i.e.,} $u^m=u^n< \dfrac{b^m_{\Tilde{k}}b^n_{\Tilde{k}}- b^m_j b^n_j}{\vert b^m_j+b^n_j-(b^m_{\Tilde{k}}+b^n_{\Tilde{k}})\vert}$ for $j\neq \Tilde{k}$, the combined opinion believes in the class that both opinions agree on, \textit{i.e.,} $\Tilde{k}=\hat{k}$.
\end{proposition}

 Define $u^m=u^n=\hat{u}$. If the beliefs of opinion $D^m$ and $D^n$ satisfy $b^m_j+b^n_j\leq b^m_{\Tilde{k}}+b^n_{\Tilde{k}}$ for all $j\neq \Tilde{k}$, the beliefs of the combined opinion $\bm{D}=\bm{D}^m \otimes \bm{D}^n$ have that,
\begin{flalign}
    b_{\Tilde{k}} &= \frac{1}{C}(b^m_{\Tilde{k}} b^n_{\Tilde{k}} + (b_{\Tilde{k}}^m + b_{\Tilde{k}}^n) \hat{u}) \nonumber\\
    &> \frac{1}{C}(b^m_j b^n_j + (b_j^m + b_j^n) \hat{u})=b_j, \quad \forall j\neq \Tilde{k}.
\end{flalign}
Therefore $b_{\Tilde{k}}$ is the largest belief in $D$, \textit{i.e.,} $\Tilde{k}=\hat{k}$.

If there exists $j\neq \Tilde{k}$ such that $b^m_j+b^n_j> b^m_{\Tilde{k}}+b^n_{\Tilde{k}}$, the beliefs of such classes satisfy,
\begin{flalign}
    b_j-b_{\Tilde{k}} &= \frac{1}{C}(b^m_j b^n_j - b^m_{\Tilde{k}} b^n_{\Tilde{k}} +  (b_j^m + b_j^n-(b_{\Tilde{k}}^m + b_{\Tilde{k}}^n))\hat{u}) \nonumber\\
    &< \frac{1}{C}(b^m_{\Tilde{k}} b^n_{\Tilde{k}} -b^m_j b^n_j + b^m_j b^n_j-b^m_{\Tilde{k}} b^n_{\Tilde{k}})=0.
\end{flalign}
Therefore $b_{\Tilde{k}}$ is also the largest belief in this case.

\begin{proposition}
 The dissimilarity between the beliefs of $\bm{D}^m$ and the combined opinion $\bm{D}$, \textit{i.e.,}  $b^m_{j} - b_{j}$, is negatively correlated with the uncertainty of opinion $\bm{D}^n$, 
 $\forall j=1,\ldots,K$ that satisfies $ b^m_{j} - b_{j}>0$.
 As $u^n$ approaches $1$, $b^m_{j} - b_{j}$ decreases to $0$.
\end{proposition}

We prove the proposition by demonstrating that the upper bound of $b^m_{j} - b_{j}$ is inversely proportional to $u_2$.
Define $b^n_m$ as the largest belief in $\bm{D}^n$, we have the following results,
\begin{flalign}\label{eq:prop2}
b^m_{j}-b_{j} & =b^m_{j}-\frac{b^m_{j} b^n_{j}+b^m_{j} u^n+b^n_{j} u^m}{\sum_{k=1}^K b_k^m b_k^n+u^n+u^m-u^m u^n} \nonumber\\
& \leq b^m_{j}-\frac{b^m_{j} u^n}{b_{m}^n+u^n+u^m-u^n u^m} \nonumber\\
& \leq b^m_{j}-\frac{b^m_{j} u^n}{1+u^m-u^n u^m}=b^m_{j} \frac{1+u^m}{\frac{1}{1-u^n}+u^m},
\end{flalign}
where the first equation is based on observation $C=\sum_{k=1}^K b_k^m b_k^n+u^n+u^m-u^m u^n$; the first inequality is tight when $b^n_{j}=0$ and $b^m_{m}=0$; the 
second inequality is based on $b^n_{m}+u^n\leq 1$.

When $u^n$ approaches $1$, $\forall j=1,\ldots,K, b^n_j=0$. The dissimilarity could be written as,
\begin{flalign}
b^m_{j}-b_{j} & =b^m_{j}-\frac{b^m_{j} b^n_{j}+b^m_{j} u^n+b^n_{j} u^m}{\sum_{k=1}^K b_k^m b_k^n+u^n+u^m-u^m u^n} \nonumber\\
&= b_j^m-\frac{b_j^m}{1+u^m-u^m}=0
\end{flalign}

\subsection{Examples of combination rules}
\begin{table}[h]
    \centering
        \resizebox{0.5\textwidth}{!}{
\begin{tabular}{|c|c|c|c|c|c|c|c|c|c|c|}
\hline 
\multirow{2}{*}{ \# } & \multicolumn{3}{|c|}{ Opinion $D^m$ } & \multicolumn{3}{|c|}{ Opinion $D^n$ } & \multirow{2}{*}{ Op } & \multicolumn{3}{|c|}{ Combined } \\
\cline{2-7}
\cline{9-11} 
\multirow{2}{*}{} & $b^m_1$ & $b^m_2$ & $u^m$ & $b^n_1$ & $b^n_2$ & $u^n$ & \multirow{2}{*}{} & $b_1$ & $b_2$ & $u$ \\
\hline 1 & 0.2 & 0.4 & 0.4 & 0.3 & 0.1 & 0.6 & $\mathrm{CBF}$ & 0.32 & 0.37 & 0.31 \\
\hline 2 & 0.1 & 0.5 & 0.4 & 0.4 & 0.2 & 0.4 & $\mathrm{BCF}$ & 0.31 & 0.49 & 0.21 \\
\hline 3 & 0.2 & 0.7 & 0.1 & 0.3 & 0.1 & 0.6 & $\mathrm{BCF}$ & 0.27 & 0.65 & 0.08 \\
\hline 4 & 0.1 & 0.2 & 0.7 & 0.2 & 0.1 & 0.7 & $\mathrm{BCF}$ & 0.24 & 0.24 & 0.52 \\
\hline
\end{tabular}
}
    \caption{Examples that illustrate the properties of the combination operators}
    \label{tab:eg_combine}
\end{table}

As illustrated in Table \ref{tab:eg_combine}, we provide several examples to demonstrate the properties of the combination operators. Example 1 illustrates that the CBF operator reduces uncertainty. Example 2 demonstrates that when two opinions exhibit equal confidence, the BCF operator outputs an opinion that believes in the consensus class (\textit{i.e.}, $\Tilde{k}=\text{arg max}_j b^m_j b^n_j$). Example 3 reveals that in the case of conflicting opinions, the BCF operator favors the confident opinion. Example 4 indicates that when both opinions are uncertain, the BCF output remains uncertain.

\clearpage
\onecolumn
\subsection{Visualization of weight for fused views in Attention-based method}

\begin{figure*}[h]
    \centering
        \scalebox{1.0}{
        \begin{tabular}{ccc}
        \fbox{\includegraphics[width=0.3\textwidth]{figs/ximenzi_1d5_s4_38_raw.pdf}} &
        \fbox{\includegraphics[width=0.3\textwidth]{figs/feilipu_3d0_s4_192_gebin_raw.pdf}} & 
        \fbox{\includegraphics[width=0.3\textwidth]{figs/feilipu_3d0_s4_231_raw.pdf}} 
        \\
        \fbox{\includegraphics[width=0.3\textwidth]{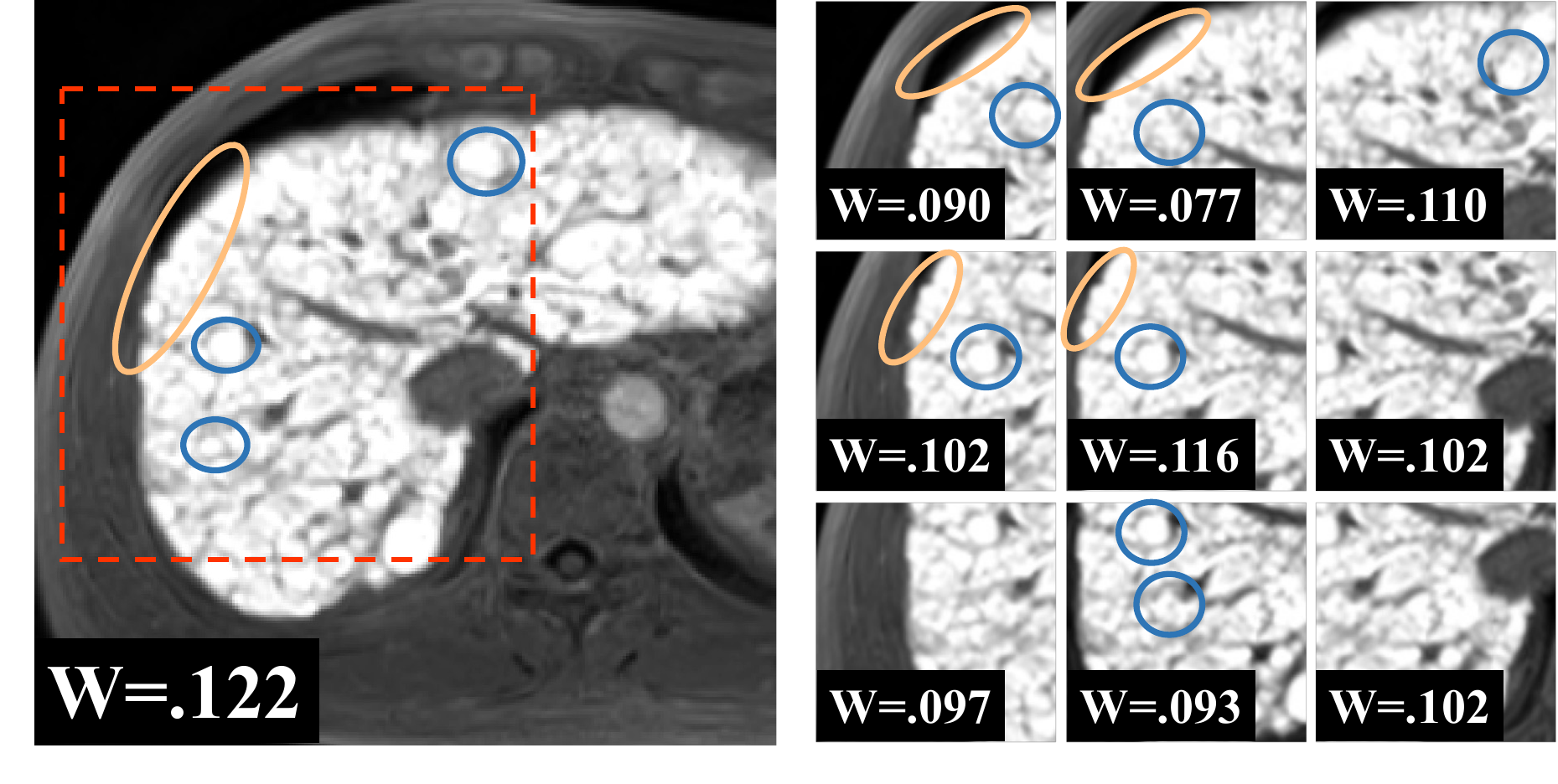}} &
        \fbox{\includegraphics[width=0.3\textwidth]{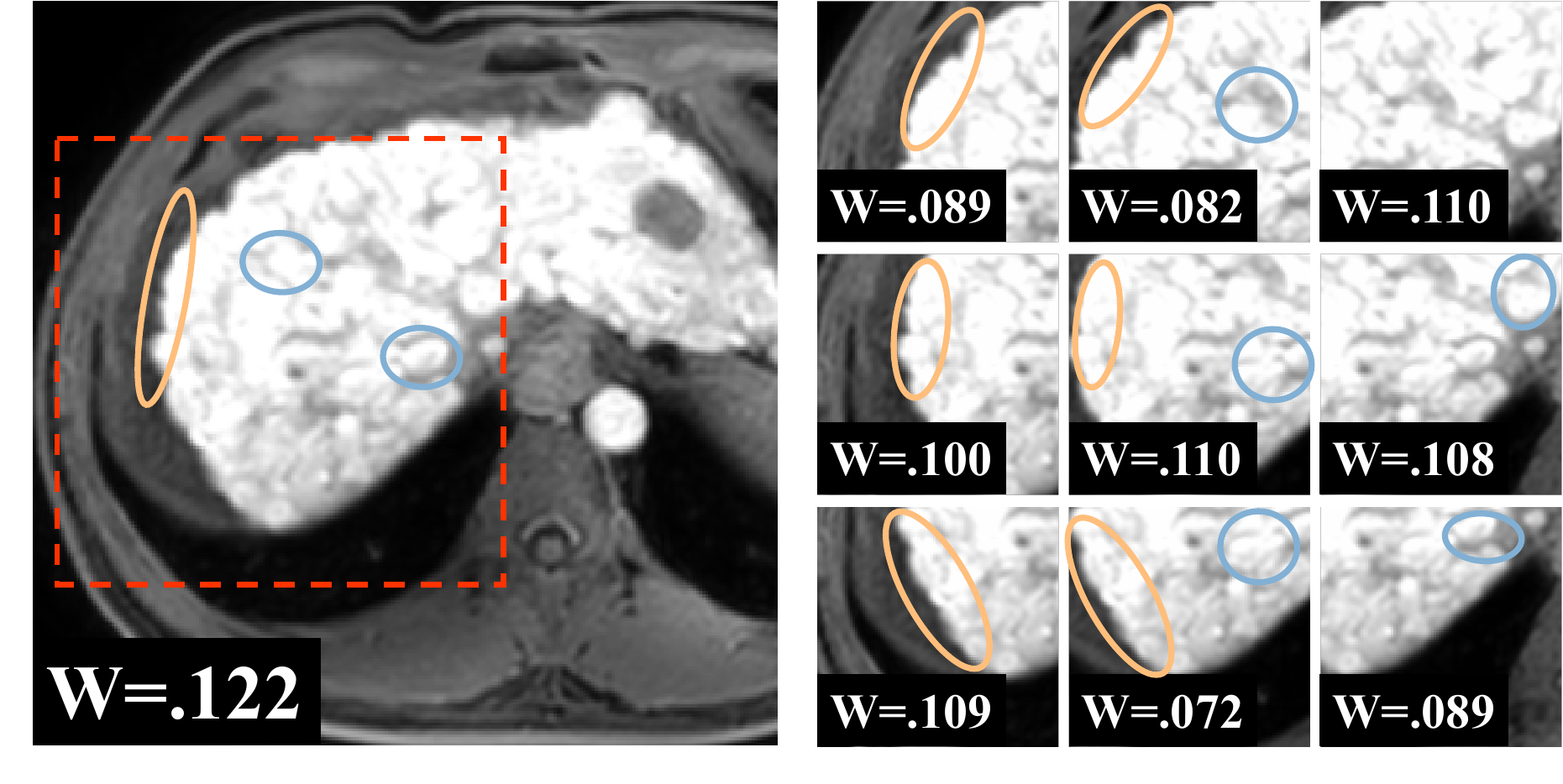}} & 
        \fbox{\includegraphics[width=0.3\textwidth]{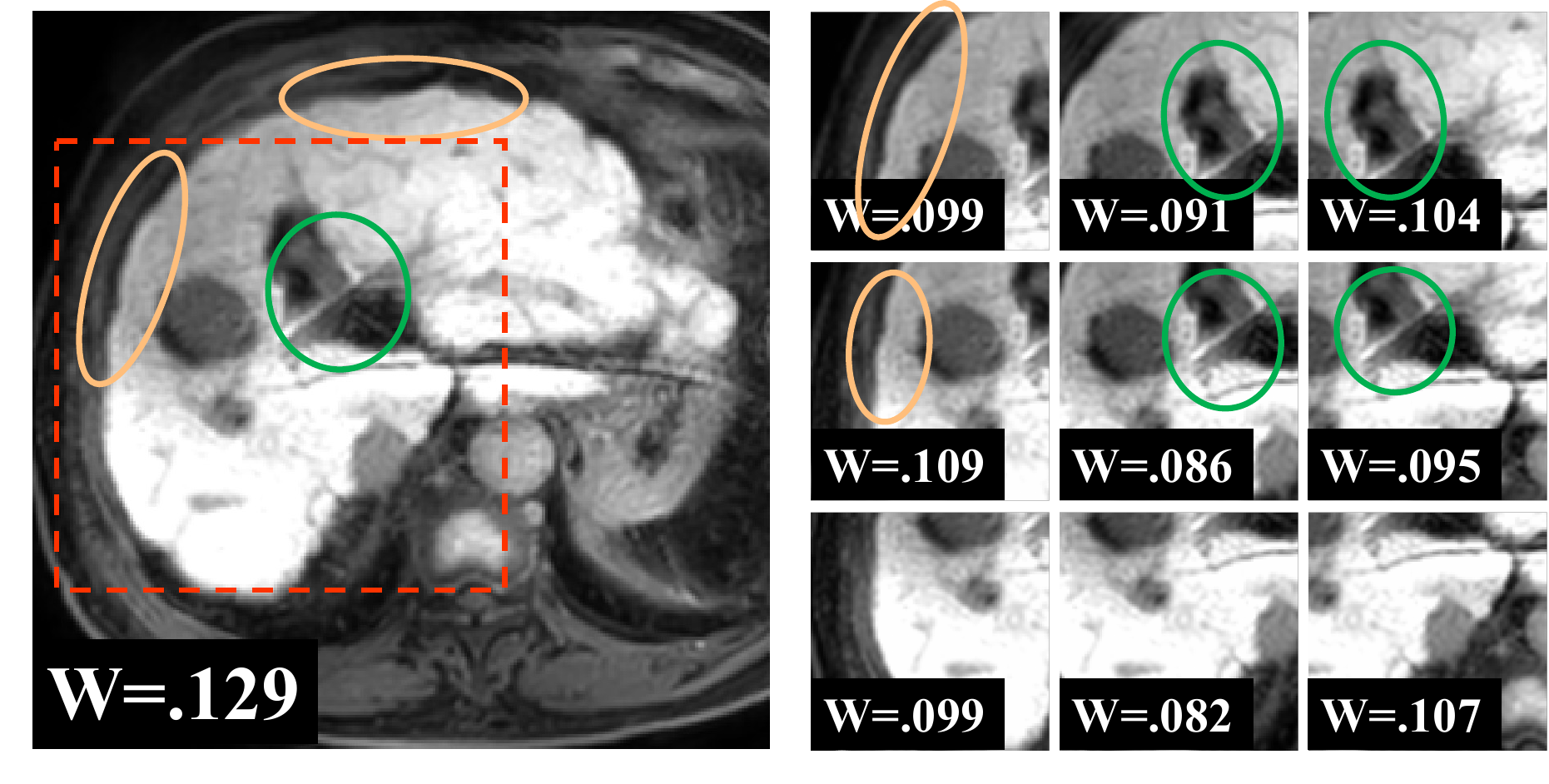}} 
        \\
        (a) & (b) & (c)
        \\
        \fbox{\includegraphics[width=0.3\textwidth]{figs/feilipu_nantong_s1_79_xuyunyun_raw.pdf}} &
        \fbox{\includegraphics[width=0.3\textwidth]{figs/ximenzi_3d0_s1_z104_limei_raw.pdf}} &
        \fbox{\includegraphics[width=0.3\textwidth]{figs/ximenzi_1d5_s1_164_raw.pdf}}  
        \\
        \fbox{\includegraphics[width=0.3\textwidth]{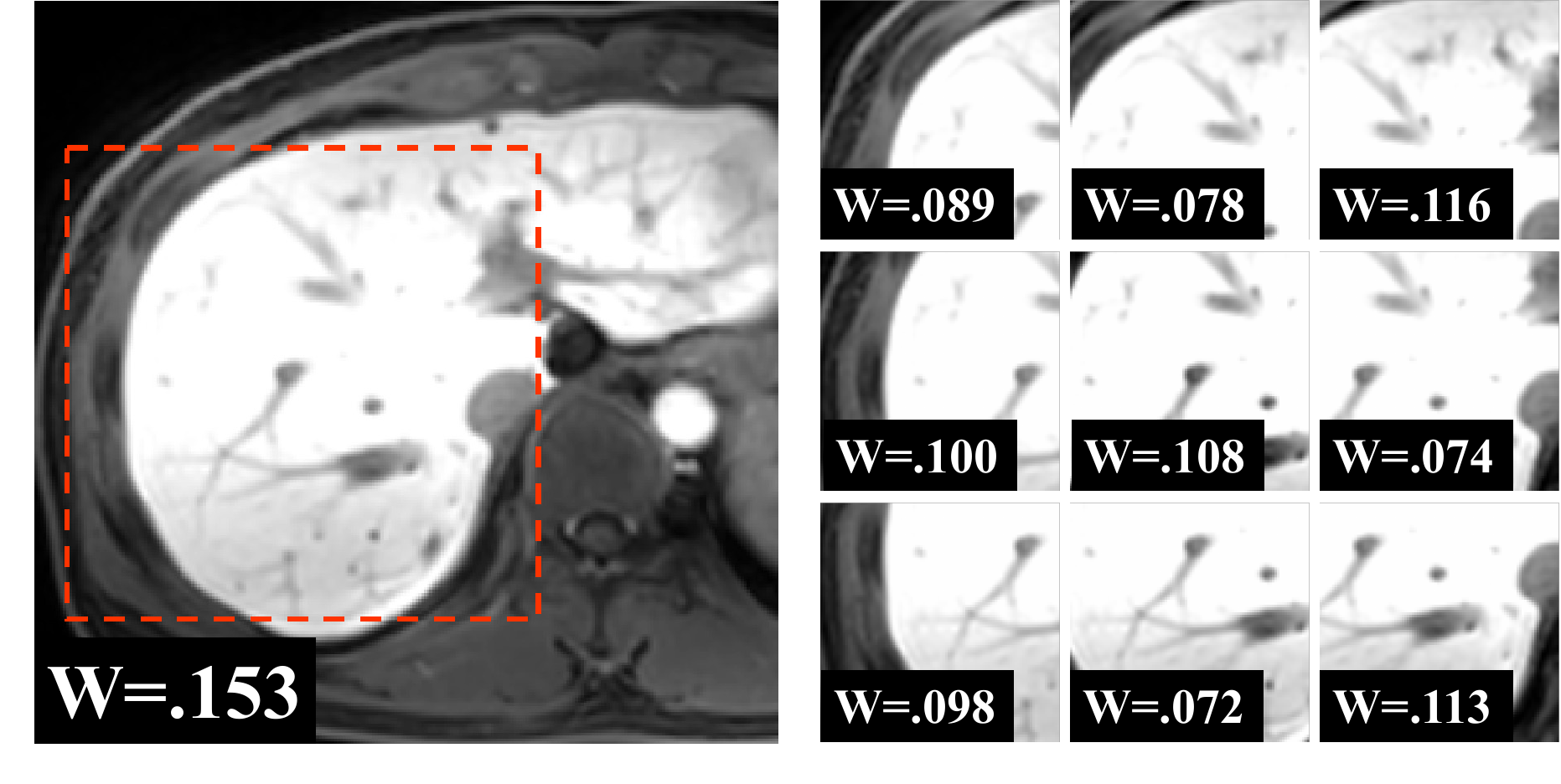}} &
        \fbox{\includegraphics[width=0.3\textwidth]{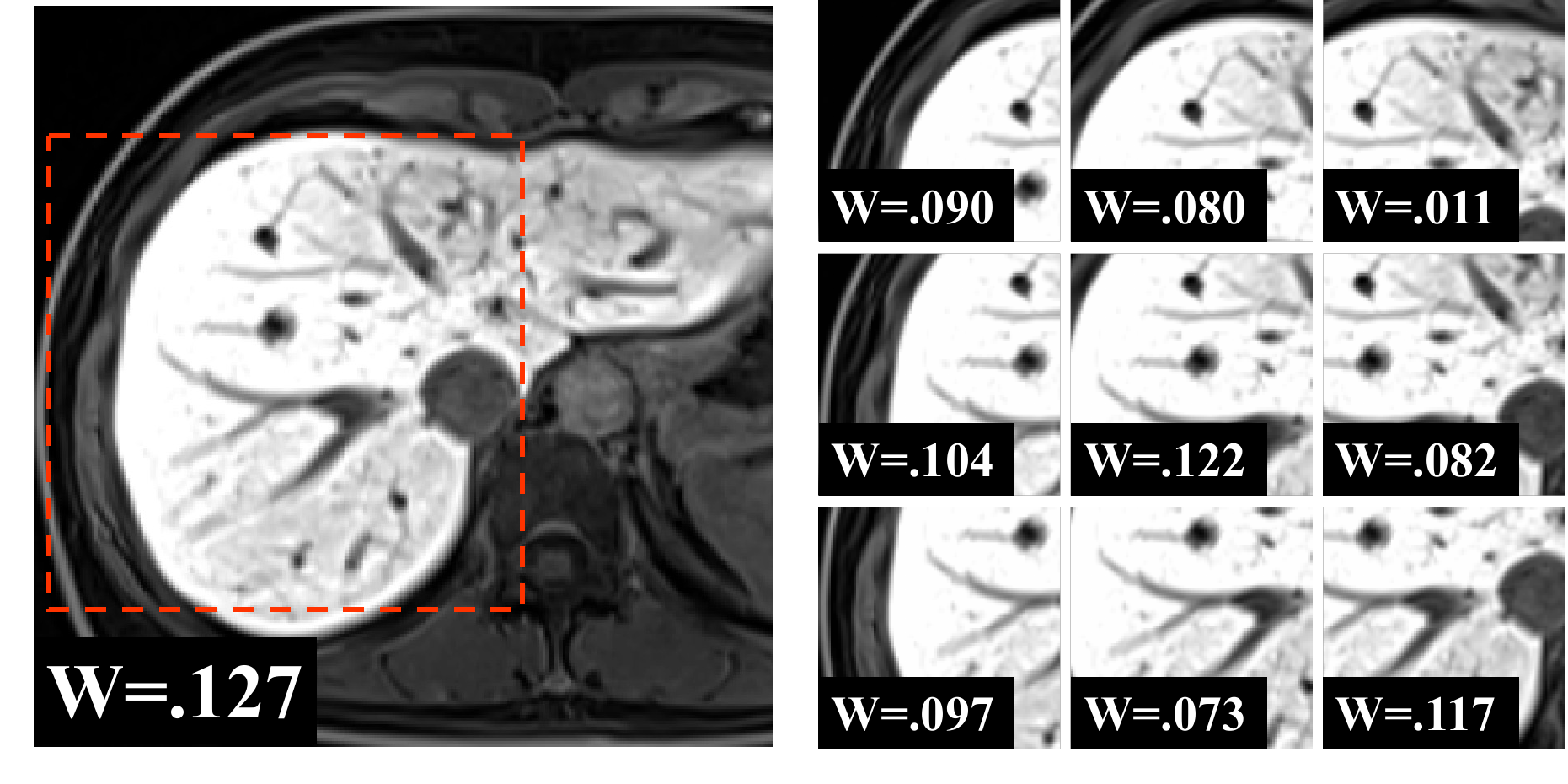}} &
        \fbox{\includegraphics[width=0.3\textwidth]{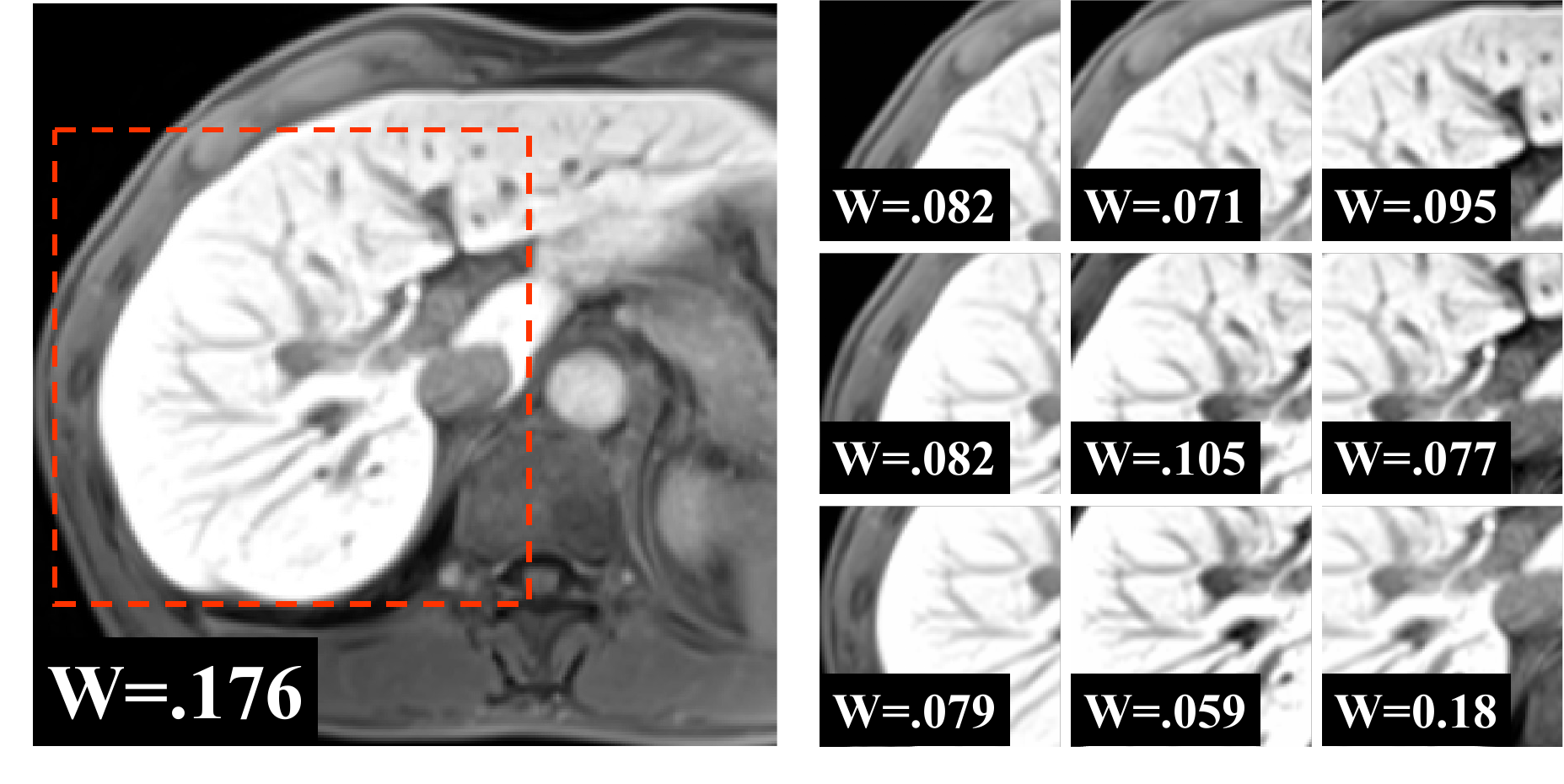}}  
        \\
     (d) & (e) & (f)
    \end{tabular}
    }
    \caption{Visualization of representative samples for stage 4 (a, b, c) and stage 1 (d, e, f). The first and third rows display the results of our method, including uncertainties (U) estimated for local and global views. The second and fourth rows correspond to the attention-based fusion method \citep{C_2018ICML_attention_fusion}, with $w$ indicating the fusion weights assigned to each view in the final layer. Our method demonstrates a more effective differentiation of views, assigning higher uncertainty to less informative views while leveraging the most useful ones. In contrast, the attention-based method assigns similar weights across views, failing to prioritize the most relevant information, which limits its ability to utilize multi-view features effectively.}
    \label{fig:weight_compare}
\end{figure*}

\clearpage
\subsection{Analysis of incorrect examples}

\begin{table*}[h]
    \centering
    \renewcommand{\arraystretch}{1.4}
    \setlength{\tabcolsep}{6pt}
        \resizebox{0.84\textwidth}{!}{
\begin{tabular}{|p{0.2\textwidth}|p{1.0\textwidth}|}
    \hline
    \textbf{Image} & \textbf{Details} \\
    \hline
    \begin{minipage}[t]{0.2\textwidth}
        \centering
        \includegraphics[width=\textwidth]{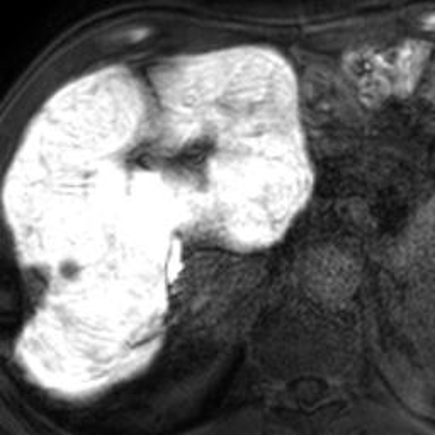}
        \vspace{2mm}
        \makecell[l]{
            \textbf{True Label:} S1 \\ 
            \textbf{Model Prediction:} \\S4 (\textit{incorrect}) \\ 
            \textbf{Clinician Prediction:} \\S4 (\textit{incorrect})
        }
    \end{minipage}
    & 
    \makecell[l]{
        \textbf{Clinician Comment (Before Knowing Ground Truth)} \\
        - Hepatic atrophy and irregular edges.\\
        - The hepatic vessels are partially obscured. \\
        \\ 
        \textbf{Clinician Analysis (After Knowing Ground Truth)} \\
        - The patient had a history of liver cancer surgery in 2014. The prior surgery resulted in liver volume reduction surface wrinkling, \\
        and morphological changes. \\
        - Respiratory artifacts have degraded image quality and obscured hepatic vascular structures. 
    } \\
    \hline
    \begin{minipage}[t]{0.2\textwidth}
        \centering
        \includegraphics[width=\textwidth]{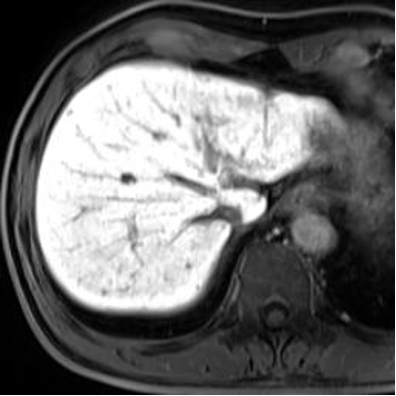}
        \vspace{2mm}
        \makecell[l]{
            \textbf{True Label:} S4 \\ 
            \textbf{Model Prediction:} \\S1-3 (\textit{incorrect}) \\ 
            \textbf{Clinician Prediction:} \\S1-3 (\textit{incorrect})
        }
    \end{minipage}
    & 
    \makecell[l]{
        \textbf{Clinician Comment (Before Knowing Ground Truth)} \\
        - The liver appears to have a normal size and shape, with
        smooth and regular contours. \\
        - The liver parenchyma demonstrates homogeneous signal
        intensity. \\
        - The hepatic vessels are clearly visible, with no signs of
        dilation or distortion. \\
        \\ 
        \textbf{Clinician Analysis (After Knowing Ground Truth)} \\
        - By reviewing the 3D liver volume reduction, diagnostic
        confidence can be increased. \\
        - After knowing the pathological results, the liver is observed to have small hardening nodules, which are not clearly visible due to\\
        the image gray value settings.
    }\\
    \hline
    \begin{minipage}[t]{0.2\textwidth}
        \centering
        \includegraphics[width=\textwidth]{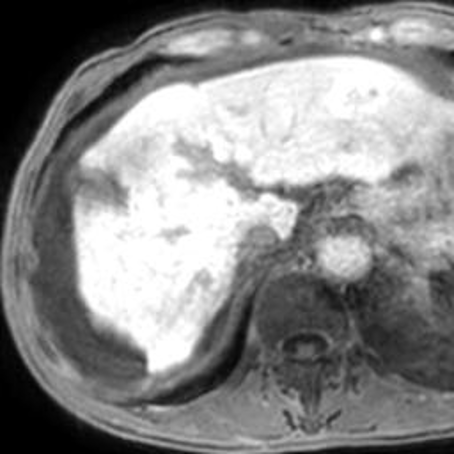}
        \vspace{2mm}
        \makecell[l]{
            \textbf{True Label:} S2 \\ 
            \textbf{Model Prediction:} \\S1-3 (\textit{incorrect}) \\ 
            \textbf{Clinician Prediction:} \\S1-3 (\textit{incorrect})
        }
    \end{minipage}
    & 
    \makecell[l]{
        \textbf{Clinician Comment (Before Knowing Ground Truth)} \\
        - Morphological changes observed, with hypertrophy of the left and caudate lobes and atrophy of the right lobe.\\
        - Parenchyma appears heterogeneous, with diffuse hepatic cirrhotic nodules visible.\\
        - Hepatic vessels are partially obscured.\\
        \textbf{Clinician Analysis (After Knowing Ground Truth)} \\
        - History of splenectomy and abdominal fluid accumulation exerting pressure on the liver, leading to localized shape changes and \\
        increased misdiagnosis rates.
    }\\
    \hline
    \begin{minipage}[t]{0.2\textwidth}
        \centering
        \includegraphics[width=\textwidth]{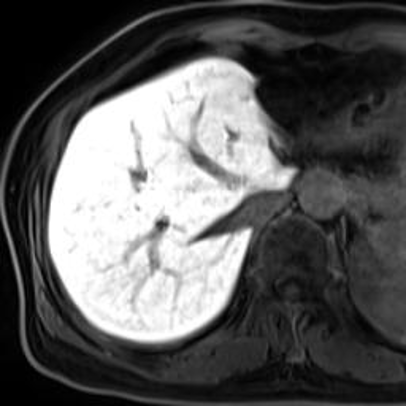}
        \vspace{2mm}
        \makecell[l]{
            \textbf{True Label:} S4 \\ 
            \textbf{Model Prediction:} \\S1-3 (\textit{incorrect}) \\ 
            \textbf{Clinician Prediction:} \\S1-3 (\textit{incorrect})
        }
    \end{minipage}
    & 
    \makecell[l]{
        \textbf{Clinician Comment (Before Knowing Ground Truth)} \\
        - Liver appears normal in size and shape, with smooth and regular contours.\\
        - Homogeneous signal intensity across the parenchyma.\\
        - Hepatic vessels are clearly visible, with no abnormalities in size or shape.\\
        \\
        \textbf{Clinician Analysis (After Knowing Ground Truth)} \\
        - History of left lobe liver surgery resulted in compensatory growth of the right lobe, masking cirrhosis-related morphological\\
        changes.
        - Suggest incorporating prior clinical history and evaluating additional sequences for comprehensive diagnosis.
    }\\
    \hline
    \end{tabular}
    }
    \caption{The table presents examples of misclassification by the MERIT model, accompanied by corresponding clinician comments and analyses. Left Column: Displays the MRI image, true label, MERIT’s prediction, and the clinician’s prediction. Right Column: Includes the clinician’s observations before and after knowing the ground truth, along with insights derived from reviewing clinical history or other relevant factors, such as prior surgeries or imaging artifacts. These examples highlight that even expert clinicians require additional context in challenging cases, underscoring the necessity of incorporating supplementary information (e.g., clinical history or 3D imaging) to enhance diagnostic accuracy. Furthermore, addressing the impact of imaging artifacts and ensuring high-quality imaging are critical to improving both model and clinician performance.}
    \label{tab:incorrect_examples}
\end{table*}

\end{document}